\pdfoutput=1
\documentclass[journal]{IEEEtran}
\IEEEoverridecommandlockouts
\usepackage[T1]{fontenc}
\usepackage{amsmath,amssymb}
\usepackage{graphicx}
\usepackage{booktabs,multirow,array}
\usepackage{algorithm}
\usepackage{algpseudocode}
\usepackage{cite}
\usepackage{url}
\usepackage{xcolor}
\usepackage{tikz}
\newif\ifinternal
\internalfalse

\newcommand{\Dist}{\mathrm{Dist}}
\newcommand{\Fup}{\overline{F}}
\newcommand{\Flo}{\underline{F}}
\newcommand{\figfile}[2]{\IfFileExists{figs/#1}{\includegraphics[#2]{figs/#1}}{\fbox{\parbox{0.9\linewidth}{\centering missing figure: \detokenize{#1}}}}}
\newcommand{\wmfigfile}[2]{\begin{tikzpicture}\node[inner sep=0pt] (img) {\figfile{#1}{#2}};\ifinternal\node[rotate=28,scale=2.6,text=black!16] at (img.center) {Prof.\ Jun Zhang};\fi\end{tikzpicture}}

\newtheorem{proposition}{Proposition}
\newtheorem{theorem}{Theorem}
\newcommand{\EoneRecords}{530}
\newcommand{\EoneN}{498}
\newcommand{\EoneSharedRuns}{20}
\newcommand{\EoneOtherRepeats}{12}

\newcommand{\EoneShorter}{440}
\newcommand{\EoneSame}{58}
\newcommand{\EoneMeanCut}{82.5}
\newcommand{\EoneMeanCutPct}{2.44}
\newcommand{\EoneMedianCut}{41.8}
\newcommand{\EoneMeanCutShortened}{93.4}
\newcommand{\EoneMaxCut}{815.0}

\newcommand{\EoneGreedyReloads}{5.31}
\newcommand{\EoneExactReloads}{7.32}
\newcommand{\EoneGreedyDetour}{6.6}
\newcommand{\EoneExactDetour}{4.3}

\newcommand{\EoneMacoCut}{32.2}

\newcommand{\EoneGaCut}{15.6}

\newcommand{\EoneMmasCut}{75.0}

\newcommand{\SelfCases}{20000}
\newcommand{\SelfFeasible}{17844}
\newcommand{\SelfSeveralPasses}{17588}

\newcommand{\EoneLabelsMean}{$1.6\times10^{5}$}
\newcommand{\EoneLabelsMedian}{$3.2\times10^{4}$}
\newcommand{\EoneLabelsMax}{$7.3\times10^{6}$}

\newcommand{\EoneCountedOrders}{100}
\newcommand{\EoneCountedMax}{8}
\newcommand{\EonePassesMean}{5.6}
\newcommand{\EonePassesMax}{15}
\newcommand{\EoneShorterWithAnt}{435}

\newcommand{\EtwoMacoXMean}{2869.9}

\newcommand{\EtwoMacoGMean}{3081.7}

\newcommand{\EtwoMacoXGDiff}{211.8}
\newcommand{\EtwoMacoXGRel}{6.87}

\newcommand{\EtwoMacoXGPholm}{1.5\times10^{-10}}
\newcommand{\EtwoGaXGDiff}{196.7}
\newcommand{\EtwoGaXGRel}{6.40}

\newcommand{\EtwoMmasXGDiff}{293.5}
\newcommand{\EtwoMmasXGRel}{9.34}

\newcommand{\EtwoMacoGaDiff}{8.3}
\newcommand{\EtwoMacoGaRel}{0.29}

\newcommand{\EtwoMacoGaPholm}{0.326}
\newcommand{\EtwoMacoMmasDiff}{21.2}
\newcommand{\EtwoMacoMmasRel}{0.74}

\newcommand{\EtwoMacoMmasPholm}{0.019}
\newcommand{\EtwoXSpread}{29.5}
\newcommand{\EtwoGSpread}{67.2}

\newcommand{\EtwoMacoMinGain}{49.5}

\newcommand{\EtwoSeedsShorterAll}{90}
\newcommand{\EtwoLongestX}{2967.7}
\newcommand{\EtwoShortestG}{2970.7}
\newcommand{\EtwoMacoXDirect}{2754.7}
\newcommand{\EtwoMacoXDetour}{115.2}

\newcommand{\EtwoMacoGDirect}{2929.3}
\newcommand{\EtwoMacoGDetour}{152.4}

\newcommand{\EtwoMacoFbMean}{6.3}
\newcommand{\EtwoMacoFbMin}{2.9}
\newcommand{\EtwoMacoFbMax}{9.4}

\newcommand{\EtwoGaFbMean}{6.0}

\newcommand{\EtwoMmasFbMean}{8.0}

\newcommand{\EtwoMacoXGreedyMean}{4169.0}
\newcommand{\EtwoMacoXGreedyMin}{3488.3}

\newcommand{\EtwoMacoPostMean}{3049.5}
\newcommand{\EtwoMacoPostGain}{32.2}
\newcommand{\EtwoMacoJointGain}{179.6}
\newcommand{\EtwoMacoPostBest}{2970.7}

\newcommand{\EtwoMacoXOptReloads}{8.4}
\newcommand{\EtwoMacoXOptAnt}{7.3}

\newcommand{\EtwoMacoGGreedyReloads}{5.5}

\newcommand{\EtwoBestLen}{2827.3}
\newcommand{\EtwoBestTspPct}{6.9}
\newcommand{\EtwoBestReloads}{8}
\newcommand{\EtwoBestAnt}{7}
\newcommand{\EtwoBestDirect}{2723.3}
\newcommand{\EtwoBestDetour}{103.9}

\newcommand{\EtwoBestGLen}{2970.7}
\newcommand{\EtwoBestGTspPct}{12.3}
\newcommand{\EtwoBestGReloads}{5}

\newcommand{\EtwoMacoMsMin}{12}
\newcommand{\EtwoMacoMsMax}{28}

\newcommand{\EtwoMacoHoursMin}{1.0}
\newcommand{\EtwoMacoHoursMax}{2.3}

\newcommand{\EtwoComputers}{3}

\newcommand{\EtwoMacoXAtFiveKMax}{3025.2}

\newcommand{\EtwoTMacoDiffA}{364.2}
\newcommand{\EtwoTMacoRelA}{11.00}
\newcommand{\EtwoTGaDiffA}{291.9}
\newcommand{\EtwoTGaRelA}{8.94}

\newcommand{\EtwoTMacoGaPholmA}{0.042}

\newcommand{\EtwoTMacoGaRelA}{0.87}
\newcommand{\EtwoTMacoDiffB}{430.5}
\newcommand{\EtwoTMacoRelB}{12.47}
\newcommand{\EtwoTGaDiffB}{377.2}
\newcommand{\EtwoTGaRelB}{11.04}

\newcommand{\EtwoTMacoGaPholmB}{0.162}

\newcommand{\EtwoTMacoGaRelB}{0.60}

\newcommand{\EtwoTMacoFbA}{11.0}

\newcommand{\EtwoTMacoFbB}{18.1}

\newcommand{\EtwoTRuns}{80}
\newcommand{\EtwoTTriple}{27}
\newcommand{\EtwoTLinuxGreedy}{40}
\newcommand{\EtwoTLinuxPrefix}{8}

\newcommand{\EthreeXAMean}{2956.8}

\newcommand{\EthreeXATime}{92.7}

\newcommand{\EthreeXBMean}{2905.3}

\newcommand{\EthreeXCMean}{2869.9}

\newcommand{\EthreeXCTimeHours}{1.75}
\newcommand{\EthreeGAMean}{3034.0}

\newcommand{\EthreeGBMean}{2992.2}

\newcommand{\EthreeGCMean}{2970.7}

\newcommand{\EthreeGCTime}{5685}

\newcommand{\EthreeGCTimeHours}{1.58}
\newcommand{\EthreeCOneDiff}{100.7}
\newcommand{\EthreeCOneRel}{3.39}

\newcommand{\EthreeCOnePholm}{1.4\times10^{-10}}
\newcommand{\EthreeCOneRatio}{1.11}
\newcommand{\EthreeCOneShorter}{30}
\newcommand{\EthreeCOneCommon}{30}
\newcommand{\EthreeCTwoDiff}{86.9}

\newcommand{\EthreeCTwoPholm}{3.3\times10^{-10}}
\newcommand{\EthreeCTwoRatio}{1.29}

\newcommand{\EthreeCThreeDiff}{77.1}

\newcommand{\EthreeCThreePholm}{6.5\times10^{-7}}
\newcommand{\EthreeCThreeRatio}{1.61}

\newcommand{\EthreeCFourDiff}{65.3}

\newcommand{\EthreeCFourPholm}{2.2\times10^{-9}}
\newcommand{\EthreeCFourRatio}{0.13}

\newcommand{\EthreeCFiveDiff}{99.9}

\newcommand{\EthreeCFivePholm}{1.4\times10^{-10}}

\newcommand{\EthreeGNoGainBC}{11}

\newcommand{\EthreeGainGAB}{41.8}
\newcommand{\EthreeGainGBC}{21.5}
\newcommand{\EthreeGainXAB}{51.5}
\newcommand{\EthreeGainXBC}{35.4}

\newcommand{\EthreeEvalRatioMin}{1100}
\newcommand{\EthreeEvalRatioMax}{1600}

\newcommand{\EthreeXAGCDiff}{13.8}
\newcommand{\EthreeXAGCRel}{0.47}
\newcommand{\EthreeMoreTimeB}{29}
\newcommand{\EthreeMoreTimeA}{61}

\newcommand{\EthreeProcs}{24}

\newcommand{\EtwoCpu}{Intel Core Ultra 9 285K}
\newcommand{\EtwoCores}{24}

\newcommand{\EtwoMemGB}{128}
\newcommand{\EtwoWindows}{11}

\newcommand{\EtwoUMacoDiff}{232.3}
\newcommand{\EtwoUMacoRel}{7.52}
\newcommand{\EtwoUGaDiff}{209.2}
\newcommand{\EtwoUGaRel}{6.79}

\newcommand{\EtwoTrendPholm}{0.0047}
\newcommand{\EtwoVRelMin}{6.79}
\newcommand{\EtwoVRelMax}{12.47}
\newcommand{\EtwoUMacoGaRel}{0.41}
\newcommand{\EtwoURefMacoRel}{7.55}
\newcommand{\EtwoURefGaRel}{6.38}
\newcommand{\EtwoUFallback}{1.2}
\newcommand{\EtwoURuns}{40}
\newcommand{\EtwoVRuns}{120}
\newcommand{\EtwoULinuxGreedy}{20}
\newcommand{\EtwoULinuxPrefix}{4}
\newcommand{\EtwoVOptima}{60}

\newcommand{\ExSeed}{209}
\newcommand{\ExGreedy}{3188.5}
\newcommand{\ExExact}{3057.7}
\newcommand{\ExCut}{130.8}
\newcommand{\ExGreedyReloads}{5}
\newcommand{\ExExactReloads}{6}
\newcommand{\ExAnt}{5}

\newcommand{\BmDemand}{3961.2}
\newcommand{\BmOneStop}{2756.0}
\newcommand{\BmOneStopPct}{69.6}
\newcommand{\BmShortfall}{1205.2}
\newcommand{\BmMinReloads}{4}

\newcommand{\EthreeGTimeSpread}{2.2}

\newcommand{\EthreeGCXBRatio}{7.5}

\newcommand{\HwPCores}{8}
\newcommand{\HwECores}{16}

\begin{document}

\title{ERD-MACO: Exact Reload Decoding in Matrix-Based Ant Colony Optimization for Hierarchical Replenishment Routing}

\author{Jun Zhang, Wei-Jie Yu, Tong Qian, Eui Jeong Kwak, Sang-Woon Jeon, Hua Ding, Qijia Jiang, Hu Jin, Haewoon Nam, and Changjun Zhou%
\thanks{This work was supported in part by the National Research Foundation of Korea under Grants RS-2025-00555463 and RS-2025-25456394.}%
\thanks{J. Zhang is with Hanyang University ERICA Campus, Nankai University, and Chaoyang University of Technology (e-mail: junzhang@ieee.org). (Corresponding author.)}%
\thanks{W.-J. Yu is with Sun Yat-sen University (e-mail: yuweijie6@mail.sysu.edu.cn).}%
\thanks{T. Qian, H. Ding, and Q. Jiang are with Nankai University (e-mail: qiantongnku@163.com; dinghua\_ieee@163.com; QijiaNKU@163.com).}%
\thanks{E. J. Kwak, S.-W. Jeon, H. Jin, and H. Nam are with Hanyang University ERICA Campus (e-mail: dmlwjd0809@hanyang.ac.kr; sangwoonjeon@hanyang.ac.kr; hjin@hanyang.ac.kr; hnam@hanyang.ac.kr).}%
\thanks{C. Zhou is with Zhejiang Normal University (e-mail: zhouchangjun@zjnu.edu.cn).}%
\thanks{An earlier version of this paper (arXiv:2611.15338v1) used a fixed greedy reload policy under the name MACO. This version replaces the policy by an exact reload decoder.}%
\ifinternal\thanks{\textcolor{red}{Internal draft of the journal version (not for circulation).}}\fi}

\newcommand{\headtext}{P\MakeLowercase{rof}. J\MakeLowercase{un} Z\MakeLowercase{hang}}
\ifinternal\markboth{\headtext}{\headtext}\else\markboth{}{}\fi

\maketitle

\begin{abstract}
In hierarchical replenishment routing, a vehicle of limited capacity delivers to every city once and reloads at cities that it has already served. Existing search methods optimize the delivery order alone and fix reloads by a greedy rule. This paper proposes ERD-MACO, exact reload decoding in matrix-based ant colony optimization, in which an exact reload decoder computes, within a label budget, the optimal reload plan of each delivery order. ERD-MACO has three components. The exact reload decoder is a labeling algorithm; dominance between labels, a distance bound and a relaxation of the stop limits keep it exact and light enough for every order. The colony construction builds the orders of a whole colony with array operations. The batch local search screens 2-opt and relocation moves by distance and decodes the selected candidates in one batch. Together, the components score orders by the best route that they admit, so short reloads made in advance replace long return trips. On a 100-city benchmark built from Korean data, ERD-MACO reaches a mean route length of \EtwoMacoXMean{}~km, against \EtwoMacoGMean{}~km for the same search with greedy decoding, a significant reduction of \EtwoMacoXGRel{}\%. At equal running time, where the greedy search performs a thousand times as many evaluations, ERD-MACO is still significantly shorter, by \EthreeCOneRel{}\%. Exact decoding also shortens the routes of a genetic algorithm and of a MAX--MIN ant system, by \EtwoGaXGRel{}\% and \EtwoMmasXGRel{}\%, and the tighter the supply, the larger the gain.
\end{abstract}


\section{Introduction}
\IEEEPARstart{A}{national} logistics network has a clear hierarchy. The capital and the metropolitan cities hold large stocks: they consume goods and can also supply a passing vehicle. Small cities have small demands and little to spare. In hierarchical replenishment routing, one vehicle of limited capacity leaves the capital, delivers to every city once and returns. Its capacity is far below the total demand, so it reloads on the way, and it may reload only at the capital or at a city that it has already served. Each city offers a limited amount per stop and allows a limited number of stops, and both limits fall with the level of the city. Two decisions therefore fix a route: the order of the deliveries, and when and where the vehicle reloads. The two decisions depend on each other. A reload costs little when the vehicle passes near a large city and much when it must return from far away, and which cities are near depends on the order.

Three lines of work are related to this problem. Routing models with intermediate stops let vehicles reload or refuel on the way. Order-first split-second methods search over orderings of the customers and compute the remaining decisions of each ordering exactly. Ant colony optimization (ACO) and its matrix-based form construct many orderings at once.

\emph{Routing with intermediate stops.} Inventory routing \cite{coelho2014}, routing with intermediate facilities \cite{angelelli2002}, the green vehicle routing problem (VRP) \cite{erdogan2012} and the multi-depot VRP with inter-depot routes \cite{crevier2007} send vehicles to dedicated facilities for reloading or refueling \cite{schiffer2019}. In the one-commodity pickup-and-delivery traveling salesman problem (TSP) \cite{hernandez2004,salazar2015}, each customer either supplies or demands the commodity. In all these models the supply points are known before the route starts. In hierarchical replenishment routing, a city becomes a supply point only after its own delivery, so the set of reload points grows along the route and depends on the order.

\emph{Order-first split-second methods.} Prins \cite{prins2004} searches over giant tours and splits every tour optimally into trips; later work lets the split choose depots, vehicle types or satellites as well \cite{prins2014} and speeds it up \cite{vidal2016}. For electric vehicles, the fixed-route vehicle-charging problem inserts charging stops into a given customer sequence \cite{montoya2017}, and a labeling algorithm solves it exactly \cite{froger2019}. These methods show that an exact sub-problem inside a search over orders pays off. Their sub-problems assume depots or stations that are given in advance and open from the start, and none of them covers supply points that the route itself creates.

\emph{ACO and matrix-based evolutionary computation.} ACO builds routes with pheromone and heuristic information \cite{dorigo1996,dorigo1997,stutzle2000,dorigo2004} and is widely used for routing \cite{bell2004,yu2009} and scheduling \cite{chen2009,zhan2010}. Matrix-based evolutionary computation (MEC), proposed by J.~Zhang \cite{zhan2022}, stores a population in matrices and processes all individuals at once; matrix-based ACO \cite{li2024maco} and the matrix-based ant colony system \cite{li2024macs} apply it to the TSP. The earlier version of this work \cite{zhang2026v1} couples a matrix-based ant colony search with a fixed greedy reload policy: the vehicle reloads only when its load does not cover the next city, and then takes the smallest immediate detour. The policy never reloads in advance. The search thus ranks an order by one particular reload plan and not by the best plan that the order admits.

The runs recorded in \cite{zhang2026v1} end in \EoneN{} distinct delivery orders. For \EoneShorter{} of them the best reload plan is shorter than the greedy plan; the mean reduction over all \EoneN{} orders is \EoneMeanCut{}~km (Section~\ref{sec:exp1}). A search that treats the reloads as a fixed rule cannot see these savings, and it cannot prefer the orders for which they are largest. The finding motivates a search in which the reload decisions are optimized together with the order.

This paper proposes ERD-MACO, which places exact reload decoding (ERD) inside a matrix-based ACO (MACO). ERD-MACO keeps the delivery order as the object of the search and computes the reload plan of every order exactly, within a label budget. The contributions are as follows.
\begin{enumerate}
\item ERD-MACO optimizes when and where to reload jointly with the delivery order: an exact reload decoder computes the optimal reload plan of every delivery order within a label budget, and the orders are evaluated in batches.
\item We state the problem as a complete model with decision variables and constraints, in which the vehicle may reload in advance, and we give its two-level form. Three properties follow: loading as much as possible loses nothing, the greedy policy is a special case, and the TSP optimum is a lower bound.
\item The exact reload decoder is a labeling algorithm with a dominance rule, a distance bound and a relaxation of the stop limits. We prove that it returns the optimal reload plan, and we check the implementation against exhaustive enumeration.
\item On a 100-city benchmark built from Korean data, the decoder solves all \EoneN{} distinct orders of the earlier runs to proven optimality and shortens \EoneShorter{} of them. In 30 runs of $3\times10^5$ evaluations, ERD-MACO shortens the mean route by \EtwoMacoXGDiff{}~km (\EtwoMacoXGRel{}\%) against the same search with greedy decoding, and exact decoding shortens the routes of a genetic algorithm and of a MAX--MIN ant system by \EtwoGaXGRel{}\% and \EtwoMmasXGRel{}\%; the three differences are significant. At equal running time, with a thousand times as many evaluations for the greedy search, ERD-MACO is still \EthreeCOneDiff{}~km (\EthreeCOneRel{}\%) shorter.
\end{enumerate}

The experiments have three parts. The first part fixes the delivery orders and compares the greedy plan with the optimal plan. The second part runs three search algorithms with both decoders for the same number of evaluations and 30 seeds, and tests the differences with the Wilcoxon rank-sum test; two of the searches are also run on three instances that differ only in the supply coefficient, and the tighter the supply, the larger the gain of exact decoding. The third part gives ERD-MACO and the same search with greedy decoding, called MACO-G, the same running time, in which MACO-G performs a thousand times as many evaluations; ERD-MACO is still \EthreeCOneRel{}\% shorter. With the exact decoder, the mean route lengths of the three searches lie within \EtwoXSpread{}~km of each other, and ERD-MACO ties with the other two searches (its mean differs from theirs by \EtwoMacoGaRel{}\% and \EtwoMacoMmasRel{}\%).

Section~\ref{sec:problem} defines the problem and reviews related work. Section~\ref{sec:method} presents ERD-MACO. Section~\ref{sec:exp} reports the experiments, and Section~\ref{sec:conclusion} concludes the paper.

\section{Problem Definition and Related Work}
\label{sec:problem}
\subsection{Problem Definition}
The instance has $n$ nodes $0,\dots,n-1$. Node 0 is the capital, Seoul in our benchmark, where the vehicle starts and ends. $D_{ij}$ is the distance between nodes $i$ and $j$; it is symmetric and satisfies the triangle inequality. Node $i$ has a demand $d_i$ with $d_0=0$ and $0<d_i\le Q$ for $i\ge1$, offers at most $s_i\ge0$ units of load at each stop, and admits at most $V_i\ge1$ stops. The vehicle has capacity $Q$. In the hierarchical setting, $s_i$ and $V_i$ decrease with the level of the city (Section~\ref{sec:benchmark}).

A solution is an extended route, the sequence of stops that the vehicle makes after it leaves Seoul:
\begin{equation}
x=(x_1,\dots,x_T),\quad x_t\in\{0,\dots,n-1\},\quad a_t\in\{\mathrm{D},\mathrm{R}\},
\label{eq:route}
\end{equation}
where $a_t$ marks stop $t$ as a delivery stop (D) or a reload stop (R), and $x_0=0$ denotes the departure. The decision variables are the number of stops $T$, the nodes $x_t$ and the types $a_t$. The load $L_t$ after stop $t$ follows from them:
\begin{align}
L_0 &= \min\{Q,\ s_0\}, \label{eq:l0}\\
L_t &= \min\{Q,\ L_{t-1}-d_{x_t}+s_{x_t}\}, && a_t=\mathrm{D}, \label{eq:ld}\\
L_t &= \min\{Q,\ L_{t-1}+s_{x_t}\}, && a_t=\mathrm{R}. \label{eq:lr}
\end{align}
An extended route is feasible if it satisfies four constraints:
\begin{align}
&\textstyle\sum_{t=1}^{T}[\,a_t=\mathrm{D},\ x_t=i\,]=1, \quad i=1,\dots,n-1, \label{eq:c1}\\
&x_t\ne0,\ L_{t-1}\ge d_{x_t} \quad \text{if } a_t=\mathrm{D}, \label{eq:c2}\\
&x_t\in\{0\}\cup\{x_{t'}: t'<t,\ a_{t'}=\mathrm{D}\},\ x_t\ne x_{t-1} \quad \text{if } a_t=\mathrm{R}, \label{eq:c3}\\
&\textstyle\sum_{t=0}^{T}[\,x_t=i\,]\le V_i, \quad i=0,\dots,n-1. \label{eq:c4}
\end{align}
Here $[\,\cdot\,]$ is 1 if the condition in the brackets holds and 0 otherwise. By \eqref{eq:c1} every node other than Seoul receives exactly one delivery. By \eqref{eq:c2} no delivery stop is made at Seoul, and the load on arrival covers the demand. By \eqref{eq:c3} the vehicle reloads only at Seoul or at a node that it has already delivered to, and not at the node where it stands. By \eqref{eq:c4} the stops at a node stay within its stop limit; the delivery stop and the departure from Seoul count, the final return to Seoul does not. At every stop the vehicle loads as much as the node offers and the capacity allows, as \eqref{eq:ld} and \eqref{eq:lr} state. The objective is the length of the route,
\begin{equation}
\min_x\ \Dist(x)=\sum_{t=1}^{T}D_{x_{t-1}x_t}+D_{x_T0}.
\label{eq:obj}
\end{equation}
The model places no condition on when a reload stop is made. The vehicle may reload while its load still covers the next delivery (an anticipatory reload), and it may leave a served node and return to it for another load; each return uses one stop of \eqref{eq:c4}.

\emph{Two-level form.} The delivery stops of a feasible route, read in order, form a delivery order $\pi=(\pi_1,\dots,\pi_{n-1})$ of the nodes other than Seoul. The reload stops made before each delivery form a reload plan $\sigma=(\sigma_1,\dots,\sigma_{n-1})$, where $\sigma_j$ is the sequence of reload nodes visited between delivery $j-1$ (or the departure) and delivery $j$. A reload stop after the last delivery cannot shorten the route, by the triangle inequality, and is excluded. A pair $(\pi,\sigma)$ determines the route, so the model is equivalent to
\begin{equation}
\min_{\pi}\ F(\pi),\qquad F(\pi)=\min_{\sigma\in\mathcal{P}(\pi)}\Dist(\pi,\sigma),
\label{eq:twolevel}
\end{equation}
where $\mathcal{P}(\pi)$ is the set of reload plans that make $\pi$ feasible and $F(\pi)=\infty$ if $\mathcal{P}(\pi)$ is empty. We call the inner problem the reload sub-problem of $\pi$. The outer problem contains the TSP: if every $s_i\ge Q$, the vehicle is full after every stop, no reload is needed, and $F(\pi)$ is the length of the closed tour of $\pi$. The problem is therefore NP-hard.

\emph{Greedy policy.} The earlier version \cite{zhang2026v1} decodes a delivery order with a fixed policy. Let the vehicle stand at node $c$ with load $L$, and let $v$ be the next node of $\pi$. While $L<d_v$, the policy makes one reload stop. The candidates are the served nodes $h\ne c$ (Seoul included) with $s_h>0$ and with a stop left under \eqref{eq:c4}. If some candidates give $\min\{Q,L+s_h\}\ge d_v$, the policy takes among them the one with the smallest detour $D_{ch}+D_{hv}-D_{cv}$; otherwise it takes the candidate with the largest ratio of the load gained to the detour plus 1~km. If there is no candidate, the policy records the shortage $d_v-L$ and continues as if the demand were covered. It then delivers to $v$ and loads there as in \eqref{eq:ld}. (The program also ends the loop after 64 reload stops before one delivery; in the \EoneRecords{} run records of \cite{zhang2026v1} the largest number is one.) The greedy fitness of $\pi$ is
\begin{equation}
f_{\mathrm{g}}(\pi)=\Dist(\pi,\sigma^{\mathrm{g}}(\pi))+M\cdot\mathrm{Short}(\pi),
\label{eq:fg}
\end{equation}
where $\sigma^{\mathrm{g}}(\pi)$ is the reload plan built by the policy, $\mathrm{Short}(\pi)$ the total recorded shortage and $M=1000$~km per unit. The policy never reloads in advance and keeps no stop in reserve, so its plan can be longer than necessary, and it can record a shortage for an order that has a feasible plan.

\emph{Properties.} Three properties of the model are used below.

\begin{proposition}[Full loading]
Suppose that the vehicle were free to load any amount between zero and the maximum at each stop. If a sequence of stops satisfies \eqref{eq:c1}--\eqref{eq:c4} for some choice of the amounts, it satisfies them when the vehicle loads the maximum at every stop, and its length is the same.
\end{proposition}
\begin{IEEEproof}
Let $L_t$ be the loads under full loading and $L'_t$ those under the other choice. Then $L'_0\le L_0$. The maps $L\mapsto\min\{Q,L-d+s\}$ and $L\mapsto\min\{Q,L+s\}$ are nondecreasing in $L$, and a vehicle that loads $y\le\min\{Q-L,s\}$ ends a stop with at most $\min\{Q,L+s\}$; by induction $L'_t\le L_t$ for all $t$. Hence \eqref{eq:c2} holds for $L_t$ whenever it holds for $L'_t$. Constraints \eqref{eq:c1}, \eqref{eq:c3}, \eqref{eq:c4} and the length do not depend on the loads.
\end{IEEEproof}

\begin{proposition}[The greedy policy as a special case]
If the greedy policy records no shortage for $\pi$, then $\sigma^{\mathrm{g}}(\pi)\in\mathcal{P}(\pi)$ and $F(\pi)\le\Dist(\pi,\sigma^{\mathrm{g}}(\pi))$.
\end{proposition}
\begin{IEEEproof}
Every reload of the policy is made at a served node other than the current one and within the stop limit, and every delivery is made with a sufficient load, so \eqref{eq:c1}--\eqref{eq:c4} hold. The inequality follows from the definition of $F$.
\end{IEEEproof}

\begin{proposition}[Lower bound]
Every feasible route is at least as long as the optimal tour of the TSP on the $n$ nodes.
\end{proposition}
\begin{IEEEproof}
Removing the reload stops of a feasible route gives a closed tour through all nodes that is not longer, by the triangle inequality.
\end{IEEEproof}

\subsection{Related Work}
\emph{Routing with intermediate stops.} Inventory routing decides delivery quantities and routes jointly \cite{coelho2014}. Routing with intermediate facilities lets vehicles reload or unload on the way \cite{angelelli2002}; the green VRP requires refueling at stations \cite{erdogan2012}; the multi-depot VRP with inter-depot routes lets vehicles reload at depots \cite{crevier2007}. Schiffer \emph{et al.} \cite{schiffer2019} review routing and location routing with intermediate stops. In these models the intermediate stops are facilities separate from the customers. Closer to our setting, the one-commodity pickup-and-delivery TSP routes one capacitated vehicle through customers that each supply or demand a single commodity \cite{hernandez2004}, and its split-demand variant allows several visits to a customer \cite{salazar2015}. In the problem of Section~\ref{sec:problem}-A, every city has a demand and becomes a reload point only after it has been served, its supply is a limit per stop that earlier stops do not use up, and the stops at a city are limited. The problem is a restricted variant of routing with intermediate stops, not a new problem class; what the models above do not provide is a method for reload points that depend on the delivery order.

\emph{Search over orders with exact sub-problems.} Order-first split-second methods separate the ordering of the customers from the other decisions. In the genetic algorithm of Prins \cite{prins2004}, a chromosome is a giant tour without trip delimiters, and the Split procedure computes the best partition of the tour into vehicle trips as a shortest path in an auxiliary acyclic graph. The review of Prins \emph{et al.} \cite{prins2014} describes extensions in which the split also selects the depot of a trip, the vehicle type, the satellites of a two-echelon system or the parking places of a truck with a trailer; with a limited heterogeneous fleet the split becomes a shortest-path problem with resource constraints. Vidal \cite{vidal2016} gives a linear-time Split for the capacitated VRP. A second group fixes the route and optimizes detours to stations. Montoya \emph{et al.} \cite{montoya2017} introduce the electric VRP with nonlinear charging and optimize the charging decisions of a fixed route inside a metaheuristic. Froger \emph{et al.} \cite{froger2019} study the fixed-route vehicle-charging problem, which decides where a given customer sequence detours to charging stations and how much it charges, and solve it exactly with a labeling algorithm; revising only the charging decisions improves 23 of 120 best known solutions. Froger \emph{et al.} \cite{froger2022} then place this labeling algorithm inside an iterated local search: every move is evaluated by solving exactly the fixed-route vehicle-charging problem of the route it changes, after a lower bound has filtered out the moves that cannot improve, and their method finds new best known solutions for 80 of 120 instances. This gain is measured against other published methods, so the effect of the exact evaluation is not separated from the differences in the search itself. Labeling algorithms with dominance are a standard method for shortest-path problems with resource constraints \cite{irnich2005}. These methods compute the secondary decisions of an order exactly, and ERD-MACO follows the same principle. Their sub-problems assume depots or stations that exist before the route starts. In the reload sub-problem of \eqref{eq:twolevel}, the reload points are the cities served so far, each with a supply per stop and a stop limit shared between its delivery and its reloads.

\emph{ACO and matrix-based evolutionary computation.} The ant system \cite{dorigo1996} uses the product of pheromone and heuristic information as the transition weight; the ant colony system (ACS) \cite{dorigo1997} adds the pseudo-random proportional rule and local and global pheromone updates; the MAX--MIN ant system (MMAS) \cite{stutzle2000} bounds the pheromone to avoid premature convergence. For vehicle routing, ACO is usually combined with candidate lists and local search \cite{bell2004,yu2009}. Genetic algorithms for the TSP are reviewed in \cite{larranaga1999}; combined with local search they are memetic algorithms \cite{moscato2003}. MEC \cite{zhan2022} stores the whole population in matrices and updates all individuals with array operations. Li \emph{et al.} apply it to ACO \cite{li2024maco} and to the ACS for the TSP \cite{li2024macs}, where the next cities of all ants and the pheromone are handled in matrix form. The earlier version of this work \cite{zhang2026v1} adds the load, service and stop-count states of all ants and decodes the orders with the greedy policy during construction.

In summary, the first line offers models but no method for reload points that the route creates; the second line offers exact sub-problems for given stations; and the third line offers a batched search whose decoder is a fixed rule. ERD-MACO combines the second and the third line for the problem of the first.

\section{ERD-MACO}
\label{sec:method}
The length of a route depends on the delivery order and on the reload plan, and the two interact. A fixed reload rule hides the interaction from the search, because every order is judged by one plan that may be far from its best plan. ERD-MACO therefore scores every order $\pi$ by $F(\pi)$, the length of its optimal reload plan, and organizes the search around this score. It has three components: the exact reload decoder, the matrix-based colony construction and the batch local search.

\begin{figure*}[t]
\centering
\figfile{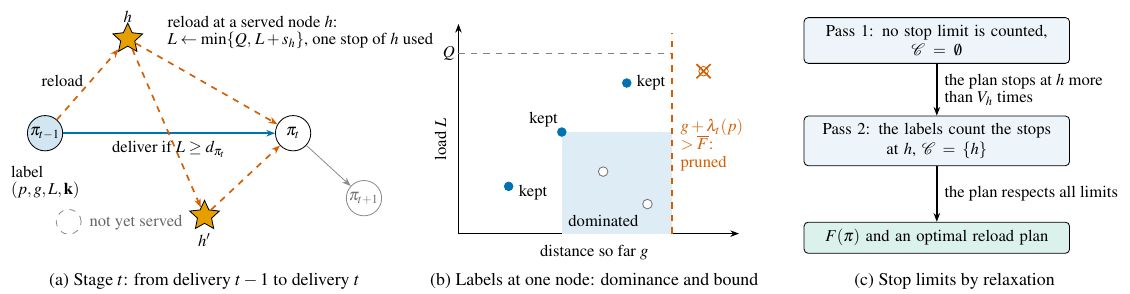}{width=0.98\textwidth}
\caption{Principle of the exact reload decoder. (a) In stage $t$ a label is extended by a delivery to $\pi_t$ or by a reload stop at a served node. (b) Among the labels of one stage at one node, a label is discarded if another label is no longer, carries no less load and has used no more stops; a label whose lower bound exceeds the bound $\Fup$ is pruned. (c) The stop limits are relaxed first and restored only at the nodes where the optimal plan of the relaxation violates them.}
\label{fig:principle}
\end{figure*}

\subsection{Exact Reload Decoder}
\label{sec:decoder}
For a fixed order $\pi$, the reload sub-problem is a shortest-path problem with resources: the vehicle moves through the deliveries of $\pi$, and its load and the stops used at the nodes restrict its moves. The decoder is a labeling algorithm \cite{irnich2005} built for this sub-problem. Fig.~\ref{fig:principle} shows its principle.

\emph{Labels.} Stage $t$ ($t=1,\dots,n-1$) covers the trip from the delivery at $\pi_{t-1}$ to the delivery at $\pi_t$, with $\pi_0=0$ for the departure. The set of served nodes of stage $t$ is $\mathcal{S}_t=\{0,\pi_1,\dots,\pi_{t-1}\}$. A label $\ell=(p,g,L,\mathbf{k})$ describes a partial route in stage $t$: the vehicle stands at node $p\in\mathcal{S}_t$, has traveled the distance $g$ and carries the load $L$. The vector $\mathbf{k}$ holds the number $k_h$ of reload stops made so far at each node $h$ of a counted set $\mathcal{C}$, which is explained below. The first label is $(0,0,\min\{Q,s_0\},\mathbf{0})$.

\emph{Extension.} A label of stage $t$ is extended in two ways (Fig.~\ref{fig:principle}(a)). If $L\ge d_{\pi_t}$, the vehicle may deliver, and the label
\begin{equation}
\bigl(\pi_t,\ g+D_{p\pi_t},\ \min\{Q,\,L-d_{\pi_t}+s_{\pi_t}\},\ \mathbf{k}\bigr)
\label{eq:ext-d}
\end{equation}
opens stage $t+1$. If $L<Q$, the vehicle may reload at any node $h\in\mathcal{S}_t\setminus\{p\}$ that has a stop left, and the label
\begin{equation}
\bigl(h,\ g+D_{ph},\ \min\{Q,\,L+s_h\},\ \mathbf{k}+\mathbf{e}_h\bigr)
\label{eq:ext-r}
\end{equation}
stays in stage $t$; $\mathbf{e}_h$ is the unit vector of $h$ if $h\in\mathcal{C}$ and zero otherwise. A node $h\in\mathcal{C}$ has a stop left if $1+k_h<V_h$, where the 1 stands for its delivery stop or, for Seoul, the departure; a node outside $\mathcal{C}$ always has one. A label with $L\ge d_{\pi_t}$ is extended in both ways; its reload extensions are the anticipatory reloads. A full vehicle is not sent to reload: once $L=Q$, the vehicle stays full until its next delivery, so reload stops made in between change only the distance.

\emph{Dominance.} Two labels of the same stage at the same node have the same served set and face the same remaining deliveries. A label $\ell'=(p,g',L',\mathbf{k}')$ dominates $\ell=(p,g,L,\mathbf{k})$ if
\begin{equation}
g'\le g,\qquad L'\ge L,\qquad \mathbf{k}'\le\mathbf{k}\ \text{(componentwise)}.
\label{eq:dom}
\end{equation}
Every continuation of $\ell$ is then feasible for $\ell'$ and not longer, so $\ell$ is discarded (Fig.~\ref{fig:principle}(b)).

\emph{Bound.} Let
\begin{equation}
\lambda_t(p)=D_{p\pi_t}+\sum_{j=t}^{n-2}D_{\pi_j\pi_{j+1}}+D_{\pi_{n-1}0}
\label{eq:lambda}
\end{equation}
be the direct length from $p$ through the remaining deliveries back to Seoul. By the triangle inequality, no completion of a label is shorter than $g+\lambda_t(p)$. Given a bound $\Fup$ on the lengths of interest, a new label is kept only if $g+\lambda_t(p)\le\Fup$ (for a label that has made the last delivery, the remaining length is $D_{p0}$).

\emph{One pass.} A pass processes the stages in order. Within a stage the labels wait in a first-in-first-out queue, and each label is extended once unless it has been dominated in the meantime. After the last stage the pass adds the return leg $D_{\pi_{n-1}0}$ to the remaining labels and returns the smallest total. Every label records its predecessor, so the pass also returns the plan of the best label. Comparisons of loads and bounds use a tolerance of $10^{-9}$, always on the side that keeps a label; the dominance test \eqref{eq:dom} uses none.

\emph{Relaxation of the stop limits.} If every label carried a stop count for every node, few labels would be comparable under \eqref{eq:dom}. The decoder therefore relaxes constraint \eqref{eq:c4}: only the nodes of the counted set $\mathcal{C}$ keep their limits, and $\mathcal{C}$ is empty at first. If the optimal plan of the relaxed problem respects all stop limits, it is optimal for the sub-problem. Otherwise the nodes whose limits it violates join $\mathcal{C}$, and a new pass follows (Fig.~\ref{fig:principle}(c)).

\emph{Stepwise bound.} The bound $\Fup$ starts close to a proven lower bound $\Flo$ of $F(\pi)$ and grows only when necessary. Initially $\Flo$ is the direct length $\lambda_1(0)$ of $\pi$, and $\Fup=\min\{F_0,\Flo+\gamma\}$ with the slack $\gamma=\gamma_0$; $F_0$ is the length of the greedy plan if the policy records no shortage, and $F_0=\infty$ otherwise. If a pass finds no plan, the relaxed problem has no plan of length $\Fup$ or less, so $\Flo$ takes the value $\Fup$ and $\gamma$ doubles. If a pass finds a plan that violates stop limits, its length is a lower bound of $F(\pi)$; $\Flo$ takes this value and $\gamma$ returns to $\gamma_0$. Once $\gamma$ exceeds 100 times the direct length, $\Fup$ is set to $F_0$. Algorithm~\ref{alg:decoder} summarizes the decoder.

\begin{algorithm}[t]
\caption{Exact reload decoder.}
\label{alg:decoder}
\small
\begin{algorithmic}[1]
\Require order $\pi$; greedy bound $F_0$; first slack $\gamma_0$; label budget $\Lambda_{\max}$.
\Ensure $F(\pi)$ and an optimal reload plan, or ``not solved''.
\State $\mathcal{C}\leftarrow\emptyset$; $\Flo\leftarrow\lambda_1(0)$; $\gamma\leftarrow\gamma_0$; $\Fup\leftarrow\min\{F_0,\Flo+\gamma\}$.
\Loop
  \State Run one pass with $\mathcal{C}$ and $\Fup$: extend the labels of the
  \Statex \hspace{\algorithmicindent} stages $1,\dots,n-1$ by \eqref{eq:ext-d} and \eqref{eq:ext-r}; discard labels
  \Statex \hspace{\algorithmicindent} by the dominance rule \eqref{eq:dom} and by the bound.
  \If{more than $\Lambda_{\max}$ labels have been offered}
    \State \Return ``not solved''.
  \EndIf
  \If{the pass finds no plan}
    \If{$\Fup=F_0$} \Return $\infty$. \EndIf
    \State $\Flo\leftarrow\Fup$; $\gamma\leftarrow2\gamma$; $\Fup\leftarrow\min\{F_0,\Flo+\gamma\}$,
    \Statex \hspace{\algorithmicindent}\hspace{\algorithmicindent} or $\Fup\leftarrow F_0$ if $\gamma>100\,\lambda_1(0)$.
  \ElsIf{the best plan violates some stop limits}
    \State Add the violated nodes to $\mathcal{C}$.
    \State $\Flo\leftarrow$ length of the plan; $\gamma\leftarrow\gamma_0$; $\Fup\leftarrow\min\{F_0,\Flo+\gamma\}$.
  \Else
    \State \Return the length of the plan and the plan.
  \EndIf
\EndLoop
\end{algorithmic}
\end{algorithm}

\begin{theorem}[Exactness]
Suppose that Algorithm~\ref{alg:decoder} ends without reaching the label budget. If it returns a plan, the plan is an optimal reload plan of $\pi$ and its length is $F(\pi)$. If it returns $\infty$ and $F_0=\infty$, then $\mathcal{P}(\pi)$ is empty.
\end{theorem}
\begin{IEEEproof}
Fix $\mathcal{C}$ and $\Fup$, and let $R$ be the relaxed sub-problem in which \eqref{eq:c4} holds for the nodes of $\mathcal{C}$ only. (i) Every label is a partial route of $R$, because the extension rules enforce \eqref{eq:c1}--\eqref{eq:c3} and the limits of $\mathcal{C}$. (ii) A label discarded by the bound has no completion of length $\Fup$ or less. (iii) Let $\ell'$ dominate $\ell$, and append any continuation of $\ell$ to $\ell'$. Both routes make the same stops from $p$ on, so they add the same distance. The load maps of \eqref{eq:ld} and \eqref{eq:lr} are nondecreasing in $L$, so the loads after $\ell'$ are not smaller than those after $\ell$, and \eqref{eq:c2} holds. The stop counts after $\ell'$ are not larger, so the limits of $\mathcal{C}$ hold. Discarding $\ell$ therefore loses no best completion, and since dominance is transitive, the same is true when $\ell'$ is discarded later. (iv) Once the vehicle is full, it stays full until its next delivery; the reload stops made in between can be removed, and the route stays feasible and does not become longer. By (i)--(iv), a pass returns the optimum of $R$ if $R$ has a plan of length $\Fup$ or less, and no plan otherwise. $R$ is a relaxation of the sub-problem, so its optimum is a lower bound of $F(\pi)$, and so is $\Fup$ after a pass without a plan; the updates of $\Flo$ are therefore valid. A returned plan respects all stop limits and attains a lower bound of $F(\pi)$, so it is optimal. If the last pass has $\Fup=F_0=\infty$ and finds no plan, $R$ has no plan, and neither has the sub-problem.
\end{IEEEproof}

The loop ends: every plan that violates stop limits adds a node to $\mathcal{C}$, and between two such additions the slack doubles until $\Fup$ reaches $F_0$. Two independent checks support the implementation. A self-test compares the decoder with the exhaustive enumeration of all reload plans on \SelfCases{} random instances of five to seven nodes, each solved with and without anticipatory reloads; \SelfFeasible{} instances are feasible, \SelfSeveralPasses{} of the 40000 solves need more than one pass, and no result differs. Every plan that the decoder returns in the experiments is replayed by a separate program, which applies \eqref{eq:l0}--\eqref{eq:c4} stop by stop and recomputes the length.

\emph{Use in the search.} ERD-MACO calls the decoder for every candidate order. To limit the cost of one call, a solve stops as ``not solved'' when it has offered $\Lambda_{\max}$ labels to the dominance test; the implementation also stops if more than 16 nodes join $\mathcal{C}$ or if the label lists of a stage exceed a memory limit; neither occurs in Experiment~1. The fitness of an order is
\begin{equation}
f(\pi)=\begin{cases}
F(\pi), & \text{if the solve succeeds and } F(\pi)<\infty,\\
f_{\mathrm{g}}(\pi), & \text{otherwise.}
\end{cases}
\label{eq:fit}
\end{equation}
A solve succeeds if it ends within the budget. In the second case the search uses the greedy fitness, which is an upper bound of $F(\pi)$ when the policy records no shortage (Proposition~2). Section~\ref{sec:exp2} reports how often this case occurs.

The decoder gives the search three things. Every order solved within the budget is ranked by the best route that it admits. Anticipatory reloads, which the greedy policy cannot make, become available. An order for which the policy records a shortage is accepted when a feasible plan exists.

\subsection{Matrix-Based Colony Construction}
ERD-MACO constructs the delivery orders of a colony of $m$ ants with array operations, in the manner of the matrix-based ACS \cite{li2024macs}. The pheromone $\boldsymbol{\tau}\in\mathbb{R}^{n\times n}$ and the heuristic $\boldsymbol{\eta}\in\mathbb{R}^{n\times n}$, with $\eta_{ij}=1/D_{ij}$ for $i\ne j$ and $\eta_{ii}=0$, are matrices. The state of the colony consists of the current nodes $\mathbf{c}\in\{0,\dots,n-1\}^m$, the unvisited mask $\mathbf{U}\in\{0,1\}^{m\times n}$ and the order matrix $\mathbf{P}\in\mathbb{Z}^{m\times(n-1)}$. At step $t$ the transition weights of all ants form the matrix
\begin{equation}
\mathbf{W}=\boldsymbol{\tau}[\mathbf{c},:]^{\circ\alpha}\circ\boldsymbol{\eta}[\mathbf{c},:]^{\circ\beta}\circ\mathbf{U}\ \in\mathbb{R}^{m\times n},
\label{eq:W}
\end{equation}
where $\boldsymbol{\tau}[\mathbf{c},:]$ stacks the rows of the current nodes, $\circ$ is the element-wise product, and $\alpha$ and $\beta$ are the pheromone and heuristic exponents. The next nodes $\mathbf{v}$ follow the pseudo-random proportional rule of the ACS: ant $k$ draws $u_k\sim U(0,1)$ and takes $v_k=\arg\max_jW_{kj}$ if $u_k<q_0$; otherwise it takes node $j$ with probability $W_{kj}/\sum_jW_{kj}$; $q_0\in[0,1]$ is the probability of taking the best node directly. Both choices are made for all rows at once. The local update
\begin{equation}
\tau_{c_kv_k}\leftarrow(1-\xi)\,\tau_{c_kv_k}+\xi\,\tau_0,\qquad k=1,\dots,m,
\label{eq:local}
\end{equation}
with the local evaporation rate $\xi\in(0,1)$, is applied to the arcs of all ants at once; an arc chosen by several ants in the same step is updated once.

After $n-1$ steps, $\mathbf{P}$ holds $m$ delivery orders, which are decoded as one batch. For every row the greedy policy gives $f_{\mathrm{g}}$ and the bound $F_0$, and Algorithm~\ref{alg:decoder} gives $F$; the fitness follows from \eqref{eq:fit}. The rows of a batch are independent of each other. The global update acts on the arcs of the best-so-far order $\pi^{\mathrm{gb}}$, including the arcs from and to Seoul: $\tau_{ij}\leftarrow(1-\rho)\tau_{ij}+\rho/f(\pi^{\mathrm{gb}})$, where $\rho\in(0,1)$ is the global evaporation rate. Here $\tau_0=1/(nf_0)$, and $f_0$ is the best fitness among 10 seed orders (randomized nearest neighbor plus 2-opt \cite{croes1958}, half of them reversed), which are decoded first. If $\pi^{\mathrm{gb}}$ does not improve for 150 iterations, $\boldsymbol{\tau}$ is reset to $\tau_0$ and the global update is applied once more.

\subsection{Batch Local Search}
Every second iteration, a batch local search improves the iteration-best order. In each pass, the gains of all 2-opt and relocation moves are computed from the distance matrix alone, with Seoul as the fixed first and last node. The 16 2-opt moves and the 16 relocation moves with the largest gains, plus 8 random moves of each type, give $K=48$ candidate orders. The candidates are decoded in one batch, and the best one is accepted if it improves $f$. The search stops when no candidate improves, after 40 passes, or when the evaluation budget is used up. The screening ignores the reloads and costs no evaluation; the decoder then judges every candidate by \eqref{eq:fit}. Every decoded order counts as one function evaluation (FE).

\subsection{Overall Framework and Complexity}
Algorithm~\ref{alg:maco} gives the framework of ERD-MACO.

\begin{algorithm}[t]
\caption{ERD-MACO.}
\label{alg:maco}
\small
\begin{algorithmic}[1]
\Require instance; $m,\alpha,\beta,q_0,\rho,\xi$; $\gamma_0$, $\Lambda_{\max}$; FE budget.
\Ensure best delivery order $\pi^{\mathrm{gb}}$ and its reload plan.
\State Decode 10 seed orders in one batch; $\pi^{\mathrm{gb}}\leftarrow$ best; $\boldsymbol{\tau}\leftarrow\tau_0\mathbf{1}$.
\While{the FE budget allows one more colony}
  \State $\mathbf{c}\leftarrow\mathbf{0}$; $\mathbf{U}\leftarrow$ all nodes except Seoul.
  \For{$t=1$ \textbf{to} $n-1$}
    \State Compute $\mathbf{W}$ by \eqref{eq:W}.
    \State Choose $\mathbf{v}$ by the pseudo-random proportional rule.
    \State Local update \eqref{eq:local} of the arcs $(\mathbf{c},\mathbf{v})$.
    \State $\mathbf{P}[:,t]\leftarrow\mathbf{v}$; update $\mathbf{U}$; $\mathbf{c}\leftarrow\mathbf{v}$.
  \EndFor
  \State Decode the $m$ rows of $\mathbf{P}$ in one batch: greedy policy,
  \Statex \hspace{\algorithmicindent} then Algorithm~\ref{alg:decoder}; fitness by \eqref{eq:fit}.
  \State Every second iteration: batch local search on the
  \Statex \hspace{\algorithmicindent} iteration-best order ($K$ candidates per batch).
  \State Update $\pi^{\mathrm{gb}}$ and the global pheromone; reset $\boldsymbol{\tau}$
  \Statex \hspace{\algorithmicindent} after 150 iterations without improvement.
\EndWhile
\State \Return $\pi^{\mathrm{gb}}$ and its reload plan.
\end{algorithmic}
\end{algorithm}

The construction of one colony takes $O(mn^2)$ operations and $O(n^2+mn)$ memory. The greedy policy adds $O(n\sum_kR_k)$ operations, where $R_k$ is the number of reload stops of ant $k$, and a local-search pass adds $O(n^2)$ operations for the screening. For the decoder we know no polynomial bound: the number of labels that do not dominate each other can grow with the number of ways to place the reloads. The budget $\Lambda_{\max}$ caps the labels of one solve, and each offered label is compared with the labels kept at one node of one stage. The memory of the decoder is the label lists of one stage and the record of predecessors. On the \EoneN{} orders of Section~\ref{sec:exp1}, a solve offers \EoneLabelsMean{} labels on average; the median is \EoneLabelsMedian{} and the maximum \EoneLabelsMax{}.

\subsection{Difference between ERD-MACO and Existing Methods}
\begin{itemize}
\item \emph{Against the earlier version with the greedy policy \cite{zhang2026v1}.} The earlier search scores an order by the plan of a fixed rule; ERD-MACO scores it by its optimal plan, which includes anticipatory reloads. By Proposition~2 the greedy plan is one element of $\mathcal{P}(\pi)$, so $F(\pi)\le f_{\mathrm{g}}(\pi)$ for every order without shortage. The load-aware transition factor of \cite{zhang2026v1}, which gave no significant gain there, is dropped.
\item \emph{Against order-first split-second methods \cite{prins2004,prins2014,vidal2016}.} ERD-MACO shares their principle: the search ranges over orders, and an exact procedure completes each order. The sub-problem differs. Split cuts the order into trips that start and end at given depots, which is a shortest path in an acyclic graph. The reload sub-problem inserts stops at nodes off the order, the nodes become available one by one as the order proceeds, and the stops at a node are limited. It needs labels with a load and stop counts.
\item \emph{Against fixed-route charging with labeling \cite{montoya2017,froger2019}.} There the stations are given, they are separate from the customers, and the amount of charge is a decision. Here the reload points are the served customers themselves, the amount is not a decision (Proposition~1), and the stop limit of a node couples its delivery with its reloads. The decoder handles these limits by relaxation and restores them only where the optimal plan needs them.
\item \emph{Against matrix-based ACO and ACS for the TSP \cite{li2024maco,li2024macs}.} In the TSP the cost of an order is the sum of its arcs. ERD-MACO keeps the matrix-based construction and adds the exact decoding of every order and the batch local search, whose candidates are decoded together.
\end{itemize}

\begin{figure*}[t]
\centering
\wmfigfile{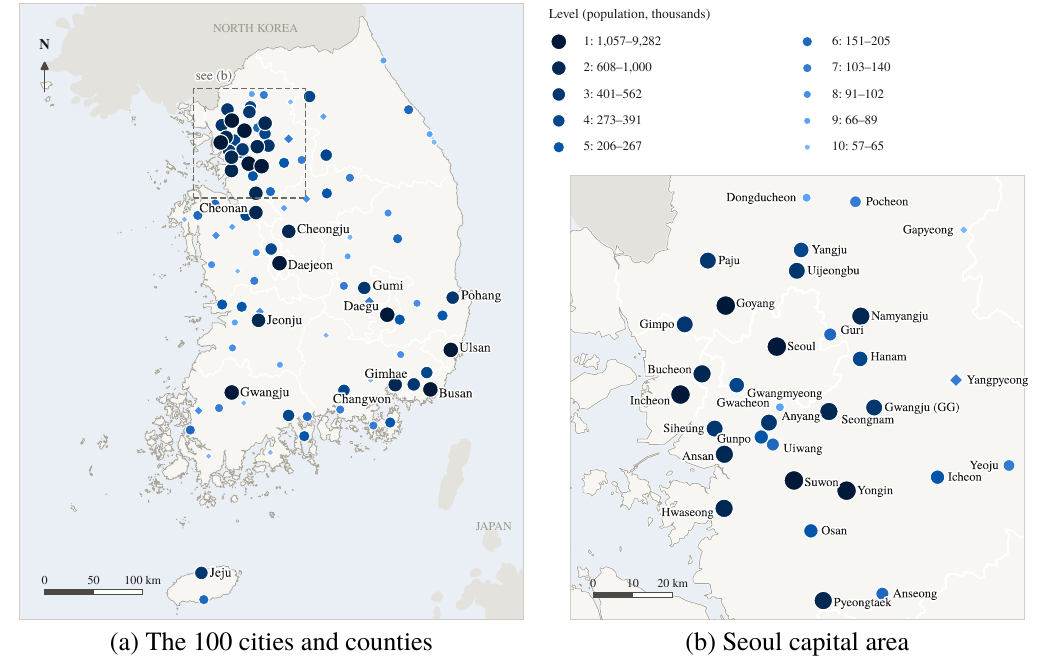}{width=0.62\textwidth}
\caption{The 100 Korean cities and counties in 10 levels by population. Darker and larger markers indicate higher levels; diamonds are counties. (a) The whole country. (b) The Seoul capital area, enlarged from the dashed box in (a).}
\label{fig:map}
\end{figure*}

\section{Experiments}
\label{sec:exp}
\subsection{Benchmark}
\label{sec:benchmark}
The benchmark uses the 100 most populous cities and counties of South Korea, with the registered population at the end of August 2026 \cite{mois2026} and the administrative boundaries of 1 July 2026 \cite{admdongkor}. Each metropolitan or special city is one node; the node Gwangju keeps the boundary that it had before its merger with Jeollanam-do on 1 July 2026. Node coordinates are those of the city hall or county office. Nodes are numbered in decreasing order of population; node 0 is Seoul, and level $z$ contains nodes $10(z-1)$ to $10z-1$, which gives 10 levels of 10 nodes (Fig.~\ref{fig:map}). $D_{ij}$ is the great-circle distance in km on a sphere of radius 6371.0088~km. The benchmark is a stylized case study and not a model of the actual distribution system: it has one commodity, one vehicle and no time windows, the two nodes on Jeju Island are connected as if by road, and demand and supply are derived from population.

Let $p_i$ be the population of node $i$ and $z_i\in\{1,\dots,10\}$ its level. Quantities are measured in units of $10^4$ residents' demand:
\begin{align}
d_i &= p_i/10^4\ (i\ge1),\qquad d_0=0, \label{eq:d}\\
s_i &= \kappa\,q^{z_i-1}\,p_i/10^4, \label{eq:s}\\
V_i &= \max\{1,\ \mathrm{round}(V_1\,r^{z_i-1})\}. \label{eq:v}
\end{align}
Here $\kappa>0$ is the supply coefficient of level 1, $q\in(0,1]$ its ratio across levels, $V_1$ the stop limit of level 1 and $r\in(0,1]$ its ratio. The supply $s_i$ is a limit per stop and is not used up by earlier stops; the stop limit $V_i$ bounds the total supply of a city. We use $Q=400$, $\kappa=1$, $q=0.6$, $V_1=5$ and $r=0.8$, so the stop limits of levels 1 to 10 are 5, 4, 3, 3, 2, 2, 1, 1, 1, 1 (Table~\ref{tab:levels}). A node of level 7 to 10 allows one stop, its delivery, and never serves as a reload point. The total demand is \BmDemand{}. The initial load and one stop at every node supply at most \BmOneStop{} units (\BmOneStopPct{}\%); the remaining \BmShortfall{} units come from reload stops of at most $Q$ units each, so every feasible route makes at least \BmMinReloads{} reload stops. The optimal TSP tour of the 100 nodes is 2646.0~km long \cite{zhang2026v1}; by Proposition~3 it is a lower bound for all routes.

\begin{table}[t]
\centering
\caption{Parameters and Statistics of the City Levels}
\label{tab:levels}
\setlength{\tabcolsep}{3pt}
\begin{tabular}{@{}ccrrrc@{}}
\toprule
Level & Population ($10^4$) & $q^{z-1}$ & Demand & Supply per stop & $V_z$\\
\midrule
1 & 928.2--105.7 & 1.000 & 1588.9 & 1988.9 & 5\\
2 & 100.0--60.8 & 0.600 & 774.3 & 464.6 & 4\\
3 & 56.2--40.1 & 0.360 & 486.6 & 175.2 & 3\\
4 & 39.1--27.3 & 0.216 & 330.3 & 71.4 & 3\\
5 & 26.7--20.6 & 0.130 & 245.3 & 31.8 & 2\\
6 & 20.5--15.1 & 0.078 & 178.8 & 13.9 & 2\\
7 & 14.0--10.3 & 0.047 & 119.1 & 5.6 & 1\\
8 & 10.2--9.1 & 0.028 & 97.3 & 2.7 & 1\\
9 & 8.9--6.6 & 0.017 & 80.5 & 1.4 & 1\\
10 & 6.5--5.7 & 0.010 & 59.9 & 0.6 & 1\\
\bottomrule
\end{tabular}\\[2pt]
\parbox{\linewidth}{\footnotesize ``Demand'' and ``Supply per stop'' are sums over the 10 nodes of a level, the latter of $\min\{Q,s_i\}$; the demand of level 1 excludes Seoul. Entries are rounded, so the demand column adds up to 3961.0 against the exact total of 3961.2.}
\end{table}

\subsection{Setup}
\begin{figure*}[t]
\centering
\figfile{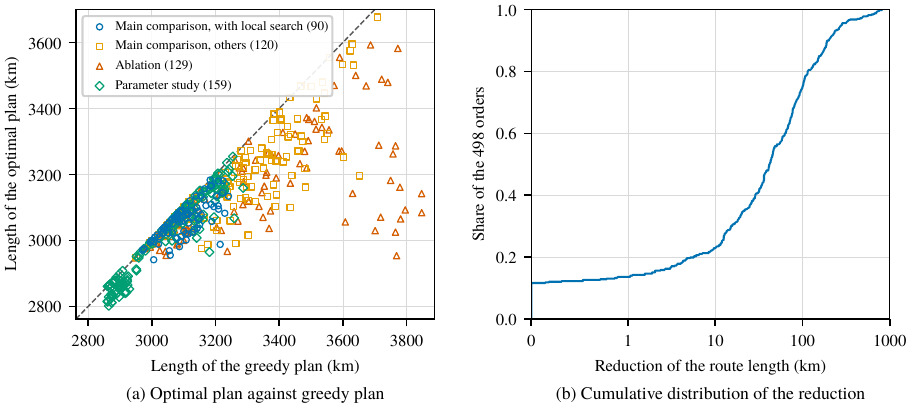}{scale=0.94}
\caption{Experiment 1. (a) Length of the greedy plan and of the optimal plan of the same delivery order, for the \EoneN{} distinct orders; points below the dashed line are orders whose optimal plan is shorter. (b) Cumulative distribution of the reduction; the horizontal axis is linear up to 1~km and logarithmic beyond.}
\label{fig:exp1}
\end{figure*}

\begin{figure*}[t]
\centering
\figfile{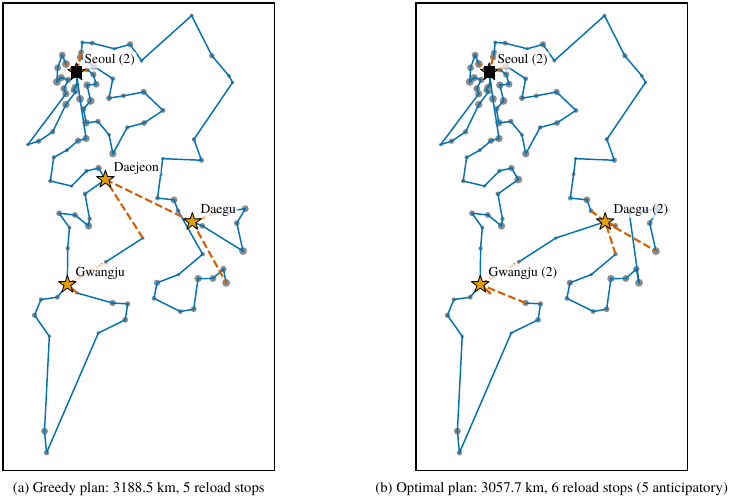}{scale=1.13}
\caption{Two reload plans of one delivery order (the final order of MACO-G with seed \ExSeed{}, which has the largest reduction among its 30 orders). Solid lines are direct legs, dashed lines are legs to or from a reload stop, stars are reload nodes (number of reload stops in parentheses), and the square is Seoul. The lines are great-circle segments, not drivable routes.}
\label{fig:example}
\end{figure*}

The experiments answer three questions. Experiment~1 asks how much shorter the optimal reload plan of a fixed order is than its greedy plan. Experiment~2 asks whether a search that scores the orders by \eqref{eq:fit} finds shorter routes than the same search with greedy decoding, for the same number of FEs. Experiment~3 asks whether this still holds when the search with greedy decoding may use the same running time.

\begin{table}[t]
\centering
\caption{Algorithms of Experiment 2 and Their Settings}
\label{tab:setup}
\setlength{\tabcolsep}{3pt}
\footnotesize
\begin{tabular}{@{}llp{5.0cm}@{}}
\toprule
Algorithm & Decoder & Search and parameters\\
\midrule
ERD-MACO & exact & Algorithm~\ref{alg:maco}; $m=50$, $\alpha=1$, $\beta=2$, $q_0=0.9$, $\rho=\xi=0.1$; local search every 2 iterations\\
MACO-G & greedy & as ERD-MACO\\
GA-LS-X & exact & genetic algorithm \cite{larranaga1999,moscato2003}: permutation encoding, population 50, binary tournament, order crossover ($p_c=0.9$), inversion ($p_m=0.3$), 2 elites, the 10 seed orders among the initial individuals; local search every 2 generations on the best individual not yet searched\\
GA-LS-G & greedy & as GA-LS-X\\
MMAS-LS-X & exact & MMAS \cite{stutzle2000}: $q_0=0$, $\rho=0.1$ on all arcs, iteration-best deposit (best-so-far every 5th iteration), $\tau_{\max}=1/(\rho f(\pi^{\mathrm{gb}}))$, $\tau_{\min}=\tau_{\max}/(2n)$, reset after 150 iterations without improvement; local search as ERD-MACO\\
MMAS-LS-G & greedy & as MMAS-LS-X\\
\bottomrule
\end{tabular}\\[2pt]
\parbox{\linewidth}{\footnotesize The exact decoder is Algorithm~\ref{alg:decoder} with $\gamma_0=5$~km and $\Lambda_{\max}=2\times10^6$. One colony and one generation both cost 50 FEs, and all searches call the local search every second iteration or generation. The FEs spent by one call differ (at most $40\times48$), and GA-LS applies it to the best individual not yet searched.}
\end{table}

Table~\ref{tab:setup} lists the algorithms of Experiment~2. The three searches are those of \cite{zhang2026v1}; MACO-G is the algorithm called MACO there. Every search runs with the greedy decoder (suffix G) and with the exact decoder (ERD-MACO, and suffix X), so each pair differs in the decoder only. All runs use $3\times10^5$ FEs and the seeds 200--229. The search parameters are those of \cite{zhang2026v1} and were fixed before its runs. The decoder parameters were fixed before the runs reported here: $\gamma_0$ during the development of the decoder on the orders of Experiment~1, and $\Lambda_{\max}$ in trial runs of 6000 FEs with seed 200, which is one of the 30 seeds.

Differences are tested with the two-sided Wilcoxon rank-sum test \cite{mann1947,derrac2011} at the 0.05 level, with the Holm correction \cite{holm1979} over five comparisons that were fixed before the runs: exact against greedy decoding for each of the three searches, and ERD-MACO against GA-LS-X and against MMAS-LS-X. Two mean route lengths that differ by at most 1\% are regarded as tied, whatever the result of the test, because a difference of this size can change when more runs are made; the tables list every difference and every $p$ value.

The comparison is repeated on three further instances. Their supply coefficients are $\kappa=1$, 0.85 and 0.72, the stop limit of level~1 is $V_1=300$ in all three, and the other parameters are unchanged, so the three instances differ only in the supply coefficient. ERD-MACO, MACO-G, GA-LS-X and GA-LS-G are run 10 times each on these instances (seeds 200--209, $3\times10^5$ FEs). All comparisons were fixed before the results were unpacked: six on the instances with $\kappa=0.85$ and 0.72 (exact against greedy decoding for the two searches on each instance, and ERD-MACO against GA-LS-X on each instance), and four more for the instance with $\kappa=1$, which was run later (exact against greedy decoding for the two searches on that instance, and, for each search, the trend of the reduction obtained with each seed as $\kappa$ decreases, with the Jonckheere--Terpstra test \cite{jonckheere1954,terpstra1952}). The Holm correction is applied within each of the two groups. All $p$ values in this paper come from the normal approximation, for the rank-sum test with tie and continuity corrections.

All programs are single-threaded C (C99) programs in which every array operation is a loop over the ants or the candidates; we compare the algorithms at equal numbers of FEs and claim no speed-up from the matrix form. The 180 runs on the benchmark instance and the \EtwoTRuns{} runs on the instances with $\kappa=0.85$ and 0.72 were made on \EtwoComputers{} computers of the same configuration (\EtwoCpu{} processor with \EtwoCores{} cores, \EtwoMemGB{}~GB of memory, Windows~\EtwoWindows{}), with 24 runs at a time on each computer; \EtwoTTriple{} of the \EtwoTRuns{} runs were made on each of the three computers and gave identical values. The program reads stored instance arrays, so its results do not depend on the mathematical library of the system. After the runs, every final order was decoded again without a label budget and its plan was replayed by the separate program; the 90 runs with greedy decoding and the first 5000 FEs of nine runs with exact decoding were repeated on a Linux computer and gave identical values. The same checks were made for the instances with $\kappa=0.85$ and 0.72, with the \EtwoTLinuxGreedy{} runs with greedy decoding and the first 5000 FEs of \EtwoTLinuxPrefix{} runs with exact decoding repeated on the Linux computer. The \EtwoURuns{} runs on the instance with $\kappa=1$ and $V_1=300$ were made on one of the three computers and were checked in the same way, with the \EtwoULinuxGreedy{} runs with greedy decoding and the first 5000 FEs of \EtwoULinuxPrefix{} runs with exact decoding repeated on the Linux computer.

\subsection{Experiment 1: Optimal Reload Plans of Fixed Orders}
\label{sec:exp1}
Experiment~1 takes the final delivery orders of the \EoneRecords{} run records of \cite{zhang2026v1}: 210 of the main comparison, 140 of the ablation and 180 of the parameter study, in which $q\in\{0.4,0.6,0.8\}$ and $r\in\{0.6,0.8,1.0\}$. \EoneSharedRuns{} records of the parameter study repeat runs of the main comparison (the cell $q=0.6$, $r=0.8$ reuses the runs of MACO-G with seeds 200--219), and \EoneOtherRepeats{} other runs end in an order that another run on the same instance also reaches. We count every order once, which leaves \EoneN{} distinct orders. All of them come from searches with greedy decoding. For each order we compare the length of the greedy plan with $F(\pi)$ from Algorithm~\ref{alg:decoder}, run without a label budget.

\begin{table}[t]
\centering
\caption{Experiment 1: Greedy and Optimal Reload Plans of the Same Delivery Orders (km)}
\label{tab:exp1}
\setlength{\tabcolsep}{3.2pt}
\begin{tabular}{@{}lrrrrrr@{}}
\toprule
Orders from & $N$ & Greedy & Optimal & Mean cut & Max cut & Shorter\\
\midrule
MACO-G & 30 & 3081.7 & 3049.5 & 32.2 & 130.8 & 21\\
GA-LS-G & 30 & 3074.9 & 3059.3 & 15.6 & 102.2 & 17\\
MMAS-LS-G & 30 & 3142.2 & 3067.1 & 75.0 & 226.7 & 30\\
Other main runs & 120 & 3305.0 & 3206.0 & 99.0 & 456.9 & 112\\
Ablation & 129 & 3290.8 & 3136.0 & 154.7 & 815.0 & 122\\
Parameter study & 159 & 3047.1 & 3012.1 & 35.0 & 216.3 & 138\\
\midrule
All & 498 & 3181.8 & 3099.3 & 82.5 & 815.0 & 440\\
\bottomrule
\end{tabular}\\[2pt]
\parbox{\linewidth}{\footnotesize $N$: number of distinct orders; an order that several runs reach is counted in the first group that contains it. ``Greedy'' and ``Optimal'': mean length of the greedy and of the optimal plan. ``Mean cut'' and ``Max cut'': mean and largest reduction. ``Shorter'': orders whose optimal plan is shorter. The first three rows are the 30 final orders of each search with local search; ``Other main runs'' are the four other configurations of the main comparison of \cite{zhang2026v1}.}
\end{table}

Table~\ref{tab:exp1} and Fig.~\ref{fig:exp1} give the results.
\begin{itemize}
\item The decoder solves all \EoneN{} orders to proven optimality. A solve needs \EonePassesMean{} passes on average and at most \EonePassesMax{}; for \EoneCountedOrders{} orders at least one stop limit has to be counted, and no order needs more than \EoneCountedMax{} counted nodes.
\item The optimal plan is shorter than the greedy plan for \EoneShorter{} orders and equal for \EoneSame{}. Over all \EoneN{} orders the mean reduction is \EoneMeanCut{}~km (on average \EoneMeanCutPct{}\% of the greedy length of an order), the median \EoneMedianCut{}~km and the maximum \EoneMaxCut{}~km; over the \EoneShorter{} shortened orders the mean is \EoneMeanCutShortened{}~km.
\item The reduction comes mainly from anticipatory reloads. The optimal plan contains at least one of them for \EoneShorterWithAnt{} of the \EoneShorter{} shortened orders and for none of the others. An optimal plan makes \EoneExactReloads{} reload stops on average, against \EoneGreedyReloads{} for the greedy plan, and the reload detours fall from \EoneGreedyDetour{}\% to \EoneExactDetour{}\% of the route length on average.
\item The reduction differs between the searches. Among the three searches with local search, the mean reduction is \EoneMacoCut{}~km for the orders of MACO-G, \EoneGaCut{}~km for those of GA-LS-G and \EoneMmasCut{}~km for those of MMAS-LS-G (Table~\ref{tab:exp1}). These orders were optimized for the greedy plan and not for $F$, so these numbers do not compare the searches; Experiment~2 does.
\end{itemize}

Fig.~\ref{fig:example} shows the mechanism on one order. The greedy plan (\ExGreedy{}~km) waits until the load is short and then returns to a large city from far away. The optimal plan (\ExExact{}~km) reloads earlier, when the vehicle is close to a large city; it makes \ExExactReloads{} reload stops instead of \ExGreedyReloads{}, \ExAnt{} of them anticipatory, and is \ExCut{}~km shorter.

\subsection{Experiment 2: Search with Exact Decoding}
\label{sec:exp2}
\begin{figure*}[t]
\centering
\figfile{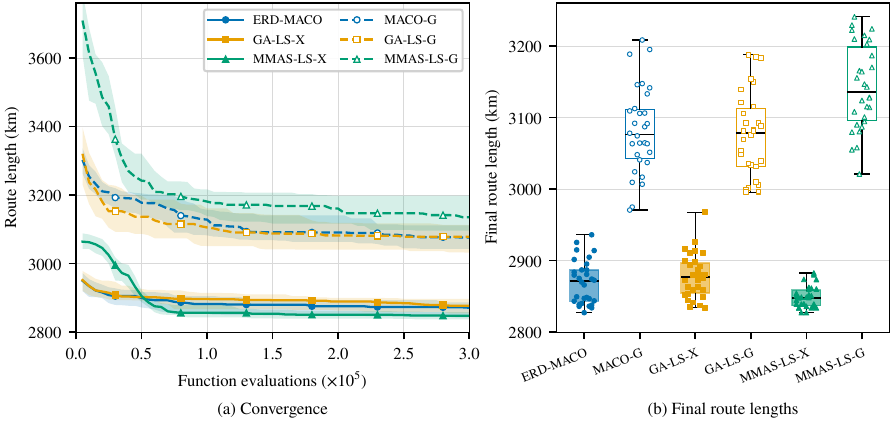}{scale=0.98}
\caption{Experiment 2. (a) Convergence: median of the 30 runs of each algorithm, with the interquartile range as a band; solid lines are searches with exact decoding, dashed lines searches with greedy decoding. (b) Final route lengths of the 30 runs; every run is a point, the boxes show the quartiles and the whiskers the range.}
\label{fig:exp2}
\end{figure*}

Tables~\ref{tab:exp2} and~\ref{tab:exp2tests} and Fig.~\ref{fig:exp2} give the results of the 180 runs on the benchmark instance; Table~\ref{tab:exp2b} gives those of the \EtwoVRuns{} runs on the three instances with $V_1=300$.

\begin{table}[t]
\centering
\caption{Experiment 2: Route Lengths of 30 Runs with $3\times10^5$ FEs (km)}
\label{tab:exp2}
\setlength{\tabcolsep}{5pt}
\begin{tabular}{@{}lrrrrr@{}}
\toprule
Algorithm & Mean & Std & Median & Best & Worst\\
\midrule
\textbf{ERD-MACO} & 2869.9 & 30.5 & 2871.3 & 2827.3 & 2936.1\\
MACO-G & 3081.7 & 60.8 & 3076.2 & 2970.7 & 3208.3\\
\addlinespace[2pt]
GA-LS-X & 2878.3 & 31.0 & 2877.3 & 2834.1 & 2967.7\\
GA-LS-G & 3074.9 & 60.0 & 3078.1 & 2995.2 & 3187.5\\
\addlinespace[2pt]
MMAS-LS-X & 2848.7 & 15.1 & 2848.2 & 2827.3 & 2882.6\\
MMAS-LS-G & 3142.2 & 59.5 & 3135.3 & 3021.0 & 3240.7\\
\bottomrule
\end{tabular}
\end{table}

\begin{table}[t]
\centering
\caption{Experiment 2: The Five Planned Comparisons (Two-Sided Wilcoxon Rank-Sum Test, Holm Correction, Level 0.05)}
\label{tab:exp2tests}
\setlength{\tabcolsep}{3.2pt}
\begin{tabular}{@{}llrrrlc@{}}
\toprule
First & Second & Diff. & Rel. & $A$ & $p_{\mathrm{Holm}}$ & \\
\midrule
ERD-MACO & MACO-G & $-211.8$ & $-6.87\%$ & 1.000 & $1.5\times10^{-10}$ & $+$\\
GA-LS-X & GA-LS-G & $-196.7$ & $-6.40\%$ & 1.000 & $1.5\times10^{-10}$ & $+$\\
MMAS-LS-X & MMAS-LS-G & $-293.5$ & $-9.34\%$ & 1.000 & $1.5\times10^{-10}$ & $+$\\
\addlinespace[2pt]
ERD-MACO & GA-LS-X & $-8.3$ & $-0.29\%$ & 0.574 & 0.326 & $=$\\
ERD-MACO & MMAS-LS-X & $+21.2$ & $+0.74\%$ & 0.304 & 0.019 & $=$\\
\bottomrule
\end{tabular}\\[2pt]
\parbox{\linewidth}{\footnotesize ``Diff.'' and ``Rel.'': mean of the first algorithm minus mean of the second, in km and relative to the second. $A$: share of the $30\times30$ pairs of runs in which the run of the first algorithm is shorter (ties count one half). $p_{\mathrm{Holm}}$: $p$ value after the Holm correction over the five tests. $+$: the first algorithm is significantly shorter. $=$: the two means differ by at most 1\% and are regarded as tied, whatever the result of the test.}
\end{table}

\begin{table}[t]
\centering
\caption{Experiment 2 on the Three Instances with Supply Coefficients 1, 0.85 and 0.72 ($V_1=300$): Route Lengths of 10 Runs with $3\times10^5$ FEs (km)}
\label{tab:exp2b}
\setlength{\tabcolsep}{3.0pt}
\begin{tabular}{@{}clrrrrrr@{}}
\toprule
$\kappa$ & Algorithm & Mean & Std & Best & Worst & Diff. & Rel.\\
\midrule
1 & \textbf{ERD-MACO} & 2858.3 & 20.5 & 2835.3 & 2887.5 & $-232.3$ & $-7.52\%$\\
 & MACO-G & 3090.7 & 62.3 & 2980.0 & 3181.0 &  & \\
 & GA-LS-X & 2870.2 & 37.3 & 2833.5 & 2955.6 & $-209.2$ & $-6.79\%$\\
 & GA-LS-G & 3079.3 & 64.6 & 2990.6 & 3182.8 &  & \\
\addlinespace[3pt]
0.85 & \textbf{ERD-MACO} & 2947.3 & 28.3 & 2911.1 & 3008.0 & $-364.2$ & $-11.00\%$\\
 & MACO-G & 3311.5 & 126.3 & 3144.3 & 3488.6 &  & \\
 & GA-LS-X & 2973.3 & 17.2 & 2939.2 & 2996.1 & $-291.9$ & $-8.94\%$\\
 & GA-LS-G & 3265.2 & 67.5 & 3183.1 & 3376.3 &  & \\
\addlinespace[3pt]
0.72 & \textbf{ERD-MACO} & 3022.5 & 29.1 & 2985.5 & 3063.0 & $-430.5$ & $-12.47\%$\\
 & MACO-G & 3453.0 & 160.7 & 3246.2 & 3702.2 &  & \\
 & GA-LS-X & 3040.8 & 21.2 & 3005.2 & 3072.9 & $-377.2$ & $-11.04\%$\\
 & GA-LS-G & 3418.0 & 100.8 & 3284.8 & 3552.7 &  & \\
\bottomrule
\end{tabular}\\[2pt]
\parbox{\linewidth}{\footnotesize ``Diff.'' and ``Rel.'': mean with exact decoding minus mean of the same search with greedy decoding, in km and relative to the latter. Each of the six differences is significant (two-sided Wilcoxon rank-sum test with the Holm correction: $p_{\mathrm{Holm}}=0.00073$ for $\kappa=1$, over the four tests planned when that instance was added, and $p_{\mathrm{Holm}}=0.0011$ for $\kappa=0.85$ and $0.72$, over the six comparisons planned for those two instances).}
\end{table}

\begin{figure*}[t]
\centering
\figfile{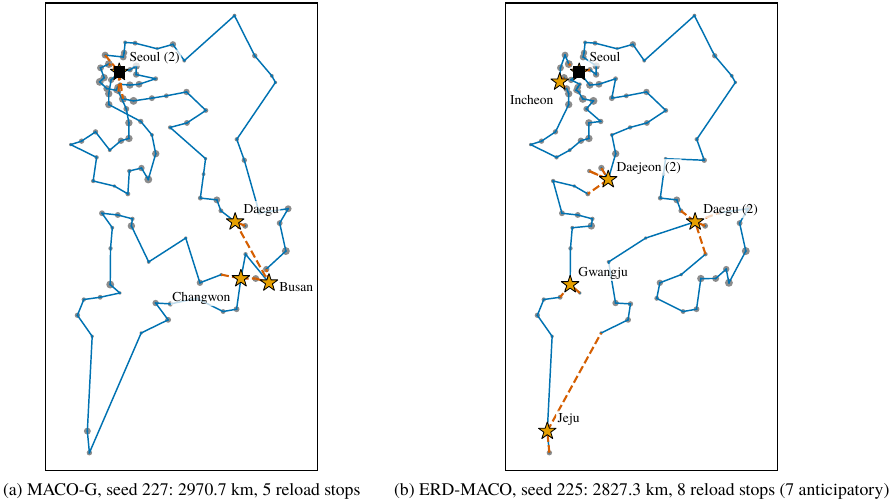}{scale=0.94}
\caption{Best routes of Experiment 2. (a) MACO-G: all \EtwoBestGReloads{} reload stops are made when the load does not cover the next city. (b) ERD-MACO: \EtwoBestAnt{} of the \EtwoBestReloads{} reload stops are anticipatory. Line styles and markers as in Fig.~\ref{fig:example}.}
\label{fig:routes}
\end{figure*}

\begin{itemize}
\item \emph{Exact against greedy decoding.} ERD-MACO reaches a mean route length of \EtwoMacoXMean{}~km, against \EtwoMacoGMean{}~km for MACO-G: exact decoding shortens the mean by \EtwoMacoXGDiff{}~km (\EtwoMacoXGRel{}\%). It shortens the mean of GA-LS by \EtwoGaXGDiff{}~km (\EtwoGaXGRel{}\%) and that of MMAS-LS by \EtwoMmasXGDiff{}~km (\EtwoMmasXGRel{}\%). The three differences are significant ($p_{\mathrm{Holm}}=\EtwoMacoXGPholm{}$ each). With the same seed, the run with exact decoding is the shorter one in all \EtwoSeedsShorterAll{} pairs, and the smallest gain of a seed is \EtwoMacoMinGain{}~km. The longest of the 90 runs with exact decoding (\EtwoLongestX{}~km) is shorter than the shortest of the 90 runs with greedy decoding (\EtwoShortestG{}~km).
\item \emph{Searching with the decoder against decoding afterwards.} Decoding the 30 final orders of MACO-G exactly, as in Experiment~1, gives \EtwoMacoPostMean{}~km on average, \EtwoMacoPostGain{}~km below MACO-G. ERD-MACO is another \EtwoMacoJointGain{}~km shorter, and each of its 30 runs ends below the best of these 30 optimal plans (\EtwoMacoPostBest{}~km). The gain therefore comes from the orders that the search selects. Under the greedy policy, the 30 final orders of ERD-MACO have plans of \EtwoMacoXGreedyMean{}~km on average and of at least \EtwoMacoXGreedyMin{}~km, so a search that ranks orders by the greedy plan places them behind all 30 final orders of MACO-G. The orders of ERD-MACO have a shorter direct length (\EtwoMacoXDirect{}~km against \EtwoMacoGDirect{}~km for MACO-G) and smaller reload detours (\EtwoMacoXDetour{}~km against \EtwoMacoGDetour{}~km). Their optimal plans make \EtwoMacoXOptReloads{} reload stops on average, \EtwoMacoXOptAnt{} of them anticipatory, against \EtwoMacoGGreedyReloads{} reload stops in the greedy plans of MACO-G.
\item \emph{Between the searches.} With greedy decoding the means of the three searches differ by up to \EtwoGSpread{}~km, with exact decoding by up to \EtwoXSpread{}~km. There ERD-MACO ties with the other two searches: its mean differs from that of GA-LS-X by \EtwoMacoGaDiff{}~km (\EtwoMacoGaRel{}\%, $p_{\mathrm{Holm}}=\EtwoMacoGaPholm{}$) and from that of MMAS-LS-X by \EtwoMacoMmasDiff{}~km (\EtwoMacoMmasRel{}\%, $p_{\mathrm{Holm}}=\EtwoMacoMmasPholm{}$), both within 1\%. These differences are far smaller than the reductions brought by exact decoding (\EtwoGaXGRel{}\% to \EtwoMmasXGRel{}\%). The search of ERD-MACO is the one of \cite{zhang2026v1} and was not adjusted to the exact decoder. The shortest route of the experiment, \EtwoBestLen{}~km, is found by one run of ERD-MACO and by two runs of MMAS-LS-X; the three routes consist of the same legs, two loops through Seoul that are traversed in either order.
\item \emph{Convergence.} After the first 5000 FEs, every run of ERD-MACO is shorter than the final mean of MACO-G (Fig.~\ref{fig:exp2}(a)); the longest of the 30 runs then measures \EtwoMacoXAtFiveKMax{}~km.
\item \emph{Fallbacks.} In ERD-MACO, \EtwoMacoFbMean{}\% of the evaluations end as ``not solved'' and use the greedy fitness (\EtwoMacoFbMin{}\% to \EtwoMacoFbMax{}\% over the runs); the share is \EtwoGaFbMean{}\% in GA-LS-X and \EtwoMmasFbMean{}\% in MMAS-LS-X. No final value is a fallback: decoded again without a label budget, all 90 final orders of the searches with exact decoding are solved to proven optimality, and the optimum equals the value reported by the search.
\item \emph{Best routes.} Fig.~\ref{fig:routes} shows the best routes of MACO-G and of ERD-MACO. The best route of MACO-G is \EtwoBestGLen{}~km long, \EtwoBestGTspPct{}\% above the TSP bound of 2646.0~km, and makes \EtwoBestGReloads{} reload stops. The best route of ERD-MACO is \EtwoBestLen{}~km long, \EtwoBestTspPct{}\% above the bound. It makes \EtwoBestReloads{} reload stops, \EtwoBestAnt{} of them anticipatory, at Seoul, Incheon, Daegu, Daejeon, Gwangju and Jeju; the detours add \EtwoBestDetour{}~km to a direct length of \EtwoBestDirect{}~km.
\item \emph{Supply coefficient.} On the three instances with $V_1=300$ (Table~\ref{tab:exp2b}), ERD-MACO is \EtwoUMacoDiff{}~km (\EtwoUMacoRel{}\%), \EtwoTMacoDiffA{}~km (\EtwoTMacoRelA{}\%) and \EtwoTMacoDiffB{}~km (\EtwoTMacoRelB{}\%) shorter than MACO-G at $\kappa=1$, 0.85 and 0.72, and exact decoding shortens the mean of GA-LS by \EtwoUGaDiff{}~km (\EtwoUGaRel{}\%), \EtwoTGaDiffA{}~km (\EtwoTGaRelA{}\%) and \EtwoTGaDiffB{}~km (\EtwoTGaRelB{}\%). The six differences are significant, and in each of the six comparisons every run with exact decoding is shorter than every run with greedy decoding. The tighter the supply, the larger the gain of exact decoding: for both searches the reduction grows as $\kappa$ decreases, and the trend is significant ($p_{\mathrm{Holm}}=\EtwoTrendPholm{}$ for each). Whether the stop limit is 5 or 300 does not change the result. At $\kappa=1$, the same 10 seeds on the benchmark instance ($V_1=5$) give reductions of \EtwoURefMacoRel{}\% and \EtwoURefGaRel{}\%, and for each of the four algorithms the means on the two instances are tied. Under exact decoding, ERD-MACO and GA-LS-X are tied on all three instances (their means differ by \EtwoUMacoGaRel{}\%, \EtwoTMacoGaRelA{}\% and \EtwoTMacoGaRelB{}\%; the last two are planned comparisons, $p_{\mathrm{Holm}}=\EtwoTMacoGaPholmA{}$ and \EtwoTMacoGaPholmB{}). The share of fallback evaluations in ERD-MACO is \EtwoUFallback{}\%, \EtwoTMacoFbA{}\% and \EtwoTMacoFbB{}\%; no final value is a fallback, and all \EtwoVOptima{} final orders of the two searches with exact decoding are solved to proven optimality.
\item \emph{Cost of exact decoding.} The comparison above is made at equal numbers of FEs. In our implementation the exact decoding of ERD-MACO takes \EtwoMacoMsMin{} to \EtwoMacoMsMax{}~ms per evaluation (mean of a run) on the computers used, so a run of ERD-MACO takes \EtwoMacoHoursMin{} to \EtwoMacoHoursMax{}~h and a run of MACO-G less than 10~s. These times describe the implementation. Experiment~3 compares ERD-MACO and MACO-G at equal running time.
\end{itemize}

\subsection{Experiment 3: Comparison at Equal Running Time}
\label{sec:exp3}
Experiment~2 compares the algorithms at equal numbers of FEs, the usual basis among evolutionary algorithms. One evaluation with the exact decoder, however, takes far longer than one with the greedy decoder, by a factor of about \EthreeEvalRatioMin{} to \EthreeEvalRatioMax{} in the runs reported here. Experiment~3 asks whether ERD-MACO still finds the shorter routes when MACO-G may use the same running time. ERD-MACO is run with $3\times10^3$, $3\times10^4$ and $3\times10^5$ FEs, and MACO-G with a thousand times as many, $3\times10^6$, $3\times10^7$ and $3\times10^8$ FEs; this gives three levels of running time, A, B and C. Each of the six groups consists of 30 independent runs on the benchmark instance with the seeds of Experiment~2. The 180 runs were made on one computer in one session, 24 at a time, and the running time of every run was recorded. The computer is one of the three computers of Experiment~2. The 30 runs of ERD-MACO with $3\times10^5$ FEs repeat those of Experiment~2 and give the same route lengths.

\begin{figure}[t]
\centering
\figfile{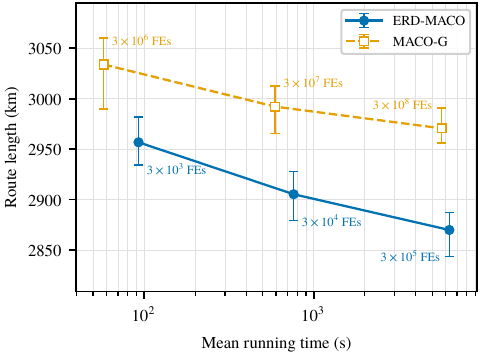}{width=\linewidth}
\caption{Experiment 3. Route length against running time. Every point is the mean of 30 runs, the bars show the interquartile range; the labels give the numbers of FEs. The time axis is logarithmic.}
\label{fig:exp3}
\end{figure}

Five comparisons and their tests were fixed before the runs were made: the two algorithms at level C (C1, the main comparison), at level B (C2) and at level A (C3); ERD-MACO at level B against MACO-G at level C (C4); and ERD-MACO at level C against the final orders of MACO-G at level C decoded once more by the exact decoder (C5). A comparison is called one at equal running time when the ratio of the two mean running times lies between 0.8 and 1.25. Tables~\ref{tab:exp3} and~\ref{tab:exp3tests} and Fig.~\ref{fig:exp3} give the results.

\begin{table}[t]
\centering
\caption{Experiment 3: ERD-MACO and MACO-G at Three Levels of Running Time, 30 Runs Each (Route Lengths in km)}
\label{tab:exp3}
\setlength{\tabcolsep}{3.2pt}
\begin{tabular}{@{}llrrrrrr@{}}
\toprule
Level & Algorithm & FEs & Time (s) & Mean & Std & Best & Worst\\
\midrule
A & \textbf{ERD-MACO} & $3\times10^3$ & 92.7 & 2956.8 & 37.2 & 2880.7 & 3025.2\\
 & MACO-G & $3\times10^6$ & 57.7 & 3034.0 & 61.7 & 2946.5 & 3195.0\\
\addlinespace[2pt]
B & \textbf{ERD-MACO} & $3\times10^4$ & 762 & 2905.3 & 31.0 & 2843.7 & 2971.4\\
 & MACO-G & $3\times10^7$ & 592 & 2992.2 & 29.6 & 2940.6 & 3044.2\\
\addlinespace[2pt]
C & \textbf{ERD-MACO} & $3\times10^5$ & 6311 & 2869.9 & 30.5 & 2827.3 & 2936.1\\
 & MACO-G & $3\times10^8$ & 5685 & 2970.7 & 25.8 & 2938.3 & 3028.3\\
\bottomrule
\end{tabular}
\end{table}

\begin{table}[t]
\centering
\caption{Experiment 3: The Five Planned Comparisons (Two-Sided Wilcoxon Rank-Sum Test, Holm Correction, Level 0.05)}
\label{tab:exp3tests}
\setlength{\tabcolsep}{2.1pt}
\begin{tabular}{@{}llrrrlcr@{}}
\toprule
ERD-MACO & MACO-G & Diff. & Rel. & $A$ & $p_{\mathrm{Holm}}$ & & Time\\
\midrule
$3\times10^5$ & $3\times10^8$ & $-100.7$ & $-3.39\%$ & 1.000 & $1.4\times10^{-10}$ & $+$ & 1.11\\
$3\times10^4$ & $3\times10^7$ & $-86.9$ & $-2.90\%$ & 0.986 & $3.3\times10^{-10}$ & $+$ & 1.29\\
$3\times10^3$ & $3\times10^6$ & $-77.1$ & $-2.54\%$ & 0.874 & $6.5\times10^{-7}$ & $+$ & 1.61\\
$3\times10^4$ & $3\times10^8$ & $-65.3$ & $-2.20\%$ & 0.958 & $2.2\times10^{-9}$ & $+$ & 0.13\\
$3\times10^5$ & $3\times10^8$$^{\dagger}$ & $-99.9$ & $-3.37\%$ & 1.000 & $1.4\times10^{-10}$ & $+$ & 1.11\\
\bottomrule
\end{tabular}\\[2pt]
\parbox{\linewidth}{\footnotesize The first two columns give the numbers of FEs of the two groups. ``Diff.'' and ``Rel.'': mean of ERD-MACO minus mean of MACO-G, in km and relative to MACO-G. $A$: share of the pairs of runs in which the run of ERD-MACO is shorter (ties count one half). $p_{\mathrm{Holm}}$: $p$ value after the Holm correction over the five tests. $+$: ERD-MACO is significantly shorter; $-$: significantly longer; $\approx$: no significant difference. ``Time'': mean running time of ERD-MACO divided by that of MACO-G; a comparison is called one at equal running time when this ratio lies between 0.8 and 1.25. $^{\dagger}$The final order of each MACO-G run is decoded once more by the exact decoder.}
\end{table}

\begin{itemize}
\item \emph{Equal running time.} At level C a run of ERD-MACO takes \EthreeXCTimeHours{}~h on average and a run of MACO-G \EthreeGCTimeHours{}~h (ratio \EthreeCOneRatio{}). In this running time ERD-MACO reaches a mean route length of \EthreeXCMean{}~km and MACO-G one of \EthreeGCMean{}~km: at equal running time the routes of ERD-MACO are still \EthreeCOneDiff{}~km (\EthreeCOneRel{}\%) shorter ($p_{\mathrm{Holm}}=\EthreeCOnePholm{}$). With the same seed, the run of ERD-MACO is the shorter one in \EthreeCOneShorter{} of the \EthreeCOneCommon{} pairs.
\item \emph{Shorter running times.} At level B the means are \EthreeXBMean{}~km for ERD-MACO and \EthreeGBMean{}~km for MACO-G, a difference of \EthreeCTwoDiff{}~km ($p_{\mathrm{Holm}}=\EthreeCTwoPholm{}$); at level A they are \EthreeXAMean{}~km and \EthreeGAMean{}~km, a difference of \EthreeCThreeDiff{}~km ($p_{\mathrm{Holm}}=\EthreeCThreePholm{}$). At these two levels the mean running time of ERD-MACO is \EthreeCTwoRatio{} and \EthreeCThreeRatio{} times that of MACO-G, outside the range fixed beforehand, so they are not comparisons at equal running time. The next comparison gives MACO-G far more time than ERD-MACO.
\item \emph{\EthreeGCXBRatio{} times the running time for MACO-G.} ERD-MACO at level B (\EthreeXBMean{}~km) against MACO-G at level C (\EthreeGCMean{}~km) differs by \EthreeCFourDiff{}~km ($p_{\mathrm{Holm}}=\EthreeCFourPholm{}$), although ERD-MACO uses only \EthreeCFourRatio{} of the running time of MACO-G. Even at level A, with a mean running time of \EthreeXATime{}~s, ERD-MACO ties with MACO-G at level C, which runs for \EthreeGCTime{}~s: the two means (\EthreeXAMean{} and \EthreeGCMean{}~km) differ by \EthreeXAGCDiff{}~km, or \EthreeXAGCRel{}\%; this pair was not among the planned comparisons and is not tested.
\item \emph{What more evaluations give.} Ten times as many FEs shorten the mean of MACO-G by \EthreeGainGAB{}~km from level A to level B and by \EthreeGainGBC{}~km from level B to level C; \EthreeGNoGainBC{} of its 30 runs do not improve at all between $3\times10^7$ and $3\times10^8$ FEs. For ERD-MACO the corresponding gains are \EthreeGainXAB{}~km and \EthreeGainXBC{}~km.
\item \emph{Decoding afterwards.} Decoding the 30 final orders of MACO-G at level C once more by the exact decoder leaves a difference of \EthreeCFiveDiff{}~km to ERD-MACO at the same level ($p_{\mathrm{Holm}}=\EthreeCFivePholm{}$): as in Experiment~2, the gain comes from searching with the exact decoder, not from decoding the final order exactly.
\end{itemize}

A hybrid search that uses the two decoders in one run was also tested, with comparisons fixed in advance; at the same budget it ties with ERD-MACO, so ERD-MACO keeps its simple form (see the supplementary material).

\section{Conclusion}
\label{sec:conclusion}
This paper proposes ERD-MACO for hierarchical replenishment routing, where a capacity-limited vehicle reloads at cities that it has already served. ERD-MACO optimizes when and where to reload jointly with the delivery order. An exact reload decoder, a labeling algorithm with dominance, a distance bound and a relaxation of the stop limits, computes the optimal reload plan of every delivery order within a label budget (on the benchmark, \EtwoMacoFbMean{}\% of the evaluations fall back to the greedy fitness); a matrix-based colony construction and a batch local search produce the orders and evaluate them in batches.

On a 100-city benchmark built from Korean data, the decoder solves all \EoneN{} distinct delivery orders of the earlier runs to proven optimality. The optimal plan is shorter than the greedy plan for \EoneShorter{} orders; the mean reduction over all \EoneN{} orders is \EoneMeanCut{}~km and the largest \EoneMaxCut{}~km, and the reduction comes mainly from anticipatory reloads. In 30 runs of $3\times10^5$ FEs, ERD-MACO reaches a mean route length of \EtwoMacoXMean{}~km, \EtwoMacoXGDiff{}~km (\EtwoMacoXGRel{}\%) below the same search with greedy decoding; decoding the orders of the greedy search exactly afterwards gives \EtwoMacoPostGain{}~km of this gain, and the rest comes from the orders that the search selects when it sees the optimal plans. At equal running time, in which the greedy search performs a thousand times as many FEs, ERD-MACO is still \EthreeCOneDiff{}~km (\EthreeCOneRel{}\%) shorter, and the difference is significant. Exact decoding shortens the routes of the genetic algorithm and of the MAX--MIN ant system by \EtwoGaXGRel{}\% and \EtwoMmasXGRel{}\%. The three differences are significant. On three instances with the same stop limit and supply coefficients of 1, 0.85 and 0.72, exact decoding shortens the routes of the two searches tested there by \EtwoVRelMin{}\% to \EtwoVRelMax{}\%, each significantly, and the tighter the supply, the larger the reduction. With exact decoding, ERD-MACO ties with GA-LS-X and with MMAS-LS-X: the means differ by less than 1\%.

The work has four limitations. First, the experiments use one stylized benchmark, in four settings of its supply and stop-limit parameters; instances with other sizes, capacities and hierarchies, generated after the algorithm is frozen, would show how far the results carry. Second, the comparison at equal running time uses one instance, one implementation and one computer on which \EthreeProcs{} runs shared the processor; on this computer, runs of MACO-G with the same FE budget differ in running time by a factor of up to \EthreeGTimeSpread{}, so the comparison rests on mean running times. The processor has \HwPCores{} performance cores and \HwECores{} efficient cores, and the runs were not bound to cores, which may explain part of this variation. The mean running times of the two algorithms lie within the range fixed beforehand only at the longest of the three levels (ratio \EthreeCOneRatio{}); at the two shorter levels ERD-MACO used \EthreeMoreTimeB{}\% and \EthreeMoreTimeA{}\% more time than MACO-G. The decoder has no polynomial bound, and a solve that reaches the label budget falls back to the greedy value; a stronger completion bound or the reuse of labels between neighboring orders in the local search would reduce the cost. Third, the search of ERD-MACO was taken from \cite{zhang2026v1} without change and ties with the other two searches under exact decoding; which pheromone rule suits exact decoding best remains open. Fourth, the local search screens its moves by distance alone; a screening that estimates the change of $F$ would spend the evaluations on better candidates.

\section*{Acknowledgment}
The original idea of this work was proposed by Professor ZHANG Jun. Under Professor Zhang's direction, an AI system, Claude (Anthropic), was used to implement the algorithms and the experiments, and help complete the manuscript. The other authors and students reviewed and proofread the code, the results and the text. The authors take full responsibility for the content.

\end{document}